\documentclass[twoside,11pt]{article}
\usepackage{blindtext}

\usepackage[preprint]{jmlr2e}

\usepackage{lastpage}

\firstpageno{1}

\usepackage[utf8]{inputenc}
\usepackage{amsmath}
\usepackage{amsfonts}
\usepackage[bb=boondox]{mathalfa}
\usepackage{xcolor}
\usepackage{nicematrix}
\usepackage{graphicx}
\usepackage{algorithm}
\usepackage{algpseudocode}
\usepackage{enumerate}
\newtheorem{assumption}{Assumption}

\DeclareMathOperator*{\tx}{\Tilde{x}}

\DeclareMathOperator*{\tr}{tr}
\DeclareMathOperator{\prox}{\mathcal{P}}
\DeclareMathOperator{\dom}{dom}
\DeclareMathOperator{\interior}{int}

\newcommand{\bb}[1]{\mathbb{#1}}

\newcommand{\td}[1]{\Tilde{#1}}

\DeclareMathOperator{\pp}{Pr}
\newcommand{\p}[1]{\pp\left[#1\right]}

\newcommand{\pover}[2]{\pp\left[#1 \mid #2\right]}

\DeclareMathOperator{\suppp}{supp}
\newcommand{\supp}[1]{\suppp(#1)}
\DeclareMathOperator{\diagg}{diag}
\newcommand{\diag}[1]{\diagg\left({#1}\right)}

\DeclareMathOperator{\kll}{KL}
\DeclareMathOperator{\tvv}{TV}
\newcommand{\kl}[2]{\kll\left({#1}\|{#2}\right)}
\newcommand{\tv}[2]{\tvv\left({#1},{#2}\right)}

\begin{document}

\title{Group-blind optimal transport to group parity \\
and its constrained variants}

\author{\name Quan Zhou \email q.zhou22@imperial.ac.uk \\
      \addr Dyson School of Design Engineering\\
      Imperial College London\\
      London SW72AZ, UK
      \AND
      \name Jakub Mare\v{c}ek \email jakub.marecek@fel.cvut.cz \\
      \addr Department of Computer Science\\
      Czech Technical University in Prague \\
      Prague 12135, the Czech Republic}
\editor{My editor}
\maketitle

\begin{abstract}
Fairness holds a pivotal role in the realm of machine learning, particularly when it comes to addressing groups categorised by protected attributes, e.g., gender, race. 
Prevailing algorithms in fair learning predominantly hinge on accessibility or estimations of these protected attributes, at least in the training process.
We design a single group-blind projection map that aligns the attribute distributions of both groups in the source data, achieving (demographic) group parity, without requiring values of the protected attribute for individual samples in the computation of the map, as well as its use.
Instead, our approach utilises the attribute distributions of the privileged and unprivileged groups in a boarder population and the essential assumption that the source data are unbiased representation of the population.
We present numerical results on synthetic data and real data.
\end{abstract}

\section{Introduction}

\textcolor{black}{
Today, machine learning plays a significant role in decision-making across nearly all facets of our lives. 
Fairness of machine learning is crucial} to avoid biased and discriminatory results, protect individuals and groups, build trust, comply with legal and ethical requirements, and ultimately improve the impact of machine learning systems \citep{mehrabi2021survey,osoba2017intelligence}.
In many applications, where machine learning underlies business practices, societal interactions, and policy-making, fairness is crucial to ensure that environmental, social and governance criteria are met \citep[ESG]{ESG2015}, which in turn promote sustainable business practises and foster positive societal impact.

It is well understood \citep{angwin2022machine,mhasawade2021machine,calders2013unbiased} that seemingly neutral machine learning models and algorithms are capable of disproportionately negative effects on certain individuals or groups based on their membership in a protected class or demographic category, even if no protected attribute is used, their use in other settings can be judged fair, and there is no explicit intention to discriminate.
\color{black}
This is particularly important in the context of foundational models, which are trained on large datasets without focusing on a single application and then adapted to new tasks using transfer learning \citep{pan2010survey,weiss2016survey}. One common approach is transductive transfer learning, also known as domain adaptation \citep{WANG2018deep,li2024comprehensive,hal2011domain}, which aims to find a good feature representation that reduces the difference between the source and target domains, as the new task may have a different data distribution from that used for training.
However, if the pre-trained model has disparate impact on different demographic groups, transferring it to a new task related to human life --- such as school admissions --- can raise ethical and legal concerns. 

\paragraph{The Challenge}

One critical concept in this context is disparate impact, which refers to a legal and social concept in which a policy, practice, or decision has a disproportionate adverse effect on a particular group, even if there is no intent to discriminate. 
For instance, even when models aim to be ``blind'' to protected attributes, they often detect patterns in correlated attributes -- such as postcode -- that act as proxies. This can lead to disparate impact, with models inadvertently discriminating against unprivileged groups based on these proxies. 
The U.S. Congress has recognised this issue by incorporating the concept of disparate impact into various anti-discrimination laws, including those related to civil rights, education, housing, and employment. 
The 80 percent rule was adopted in 1978 by the US Equal Employment Opportunity Commission. The rule serves as a guideline used in the context of disparate impact analysis, particularly in employment and civil rights law. It suggests that a selection rate for a protected group should be at least 80\% of the selection rate for the group with the highest selection rate. 
\color{black}


Generally speaking, to mitigate the disparate impact, two common strategies have been employed within the broader field of machine learning fairness.
\textcolor{black}{
} The first strategy directly adjusts the values of attributes, labels, model outputs (estimated labels) or any combination thereof.
Initially, \cite{zemel2013learning} proposed a mapping to some fair representations of attributes.
\cite{feldman2015certifying} suggested modifying attributes so that the distributions for privileged and unprivileged groups become similar to a ``median'' distribution, making it harder for the algorithm to differentiate between the two groups.
The principle has been expanded to adjusting both attributes and the label in \cite{calmon2017optimized} and further to projecting distributions of attributes to a barycentre \citep{gordaliza2019obtaining,yang2022obtaining}, which introduces the least data distortion.
\cite{oneto2020fairness,chzhen2020fair,gouic2020projection} performed post-processing of the model outputs by transporting the distributions of the outputs of each group to a barycentre. 
\cite{jiang2020identifying} corrected for a range of biases by re-weighting the training data.

The second strategy incorporates a regularisation term to penalise discriminatory effects whilst building some classification or prediction models.
\cite{quadrianto2017recycling} designed regularisation components using the maximum mean discrepancy criterion of \cite{gretton2012kernel} to encourage the similarity of the distributions of the prediction outputs across the privileged and unprivileged groups, with the assumption of protected attributes only available at training time.
\cite{jiang2020wasserstein} used Wasserstein distance of the distributions of outputs between privileged and unprivileged groups as the regularisation in logistic regression, again without any requirements on the availability of the protected attribute at test time. 
The Wassersterin regularisation is also used in neural-network classifiers \citep{risser2022tackling} and various applications \citep{jourdan2023optimal} thereof.
\cite{buyl2022optimal} design a regularisation term by the minimum cost to transport a classifier's score function to a set of fair score functions, and generalise the fairness measures allowed.


All the methods mentioned above use the protected attribute of each sample (data point) as a prerequisite for modifying the values differently for different groups (in the first strategy), or
measuring the loss of group-fairness regularisation term in iterations of training the classification or prediction models (in the second strategy).
A key challenge, which has not been explored in depth, is the unavailability of protected attributes (e.g., gender, race) in many practical applications. In practice, 
protected attributes are generally unavailable or inaccessible. In many jurisdictions, businesses cannot collect data on the race of their customers. 
\textcolor{black}{In the European Union, the AI Act will allow the collection of protected attributes in so-called sandboxes to audit the fairness of algorithms, but not to train them. Suppose a large bank with a subsidiary in France is involved in model validation. While the AI Act (Regulation (EU) 2024/1689, which lays down harmonised rules on artificial intelligence) provides legal provisions (Article 57 on AI regulatory sandboxes and subsequent articles) for collecting sensitive personal information to estimate bias in AI systems, the French ethos generally discourages collecting data on religion, sex, or ethnic origin. Even if the bank decides to collect sensitive personal information from, say, a hundred thousand customers, the return rate for such a questionnaire may be low, and response rates could vary significantly across different groups.
Similarly in the U.S.A., CFBP Consumer Law and Regulations, 12 CFR 1002.5 states, ``Creditors may not request or collect information about an applicant’s race, color, religion, national origin, or sex. Exceptions to this rule generally involve situations where the information is necessary to test for compliance with fair lending rules''.
}

\paragraph{Contributions}

We introduce novel methods for removing disparate impact in transfer learning (also related to domain adaptation and fairness repair), without requiring the protected attribute to be revealed.
Our methods are related to the ``total repair'' and ``partial repair'' schemes of \cite{feldman2015certifying} and  \cite{gordaliza2019obtaining}, which project the distributions of attributes for privileged and unprivileged groups to a target distribution interpolating between the groups.
However, the previous schemes \citep{gordaliza2019obtaining,feldman2015certifying} require the protected attribute to define groups, while the protected attribute is either simply unavailable, or even illegal to collect
in many settings. 

Our work follows the idea \citep[Slide 23]{Quinn2023} to use marginals from another data set in transfer learning towards a data set whose bias is, at least in part, determined by an unavailable protected attribute. 
In the setting envisioned by the AI Act in the European Union, this additional data set could be collected within so-called regulatory sandbox \cite{allen2019regulatory,morgan2023anticipatory}, wherein a regulator and the developer of an AI system agree to override data privacy protection
in a well-defined fashion.
In many other settings, the data could be collected in the census.
When no such data exists, such as when France bars the collection of data on the race and ethnicity of its citizens even in the census \citep{Ndiaye2020},
or India bars the collection of data on caste status in its census \citep{Bose2023}, 
it would seem difficult to address the bias, indeed.

We develop algorithms to realise this objective.  
The membership of each sample is used to calculate distinct projections for various groups and to modify the value of attributes via group-wise projections.
We extend these schemes to modifying the values of attributes via one group-blind projection map, and achieve equalised distributions of modified attributes between privileged and unprivileged groups.
The sample membership is not required for computing such group-blind projection map, or for modifying the values of attributes via this group-blind projection map.
We require only the population-level information regarding attribute distributions for both privileged and unprivileged groups, as well as the assumptions that source data are collected via unbiased sampling from the population.
The target distribution in our framework is not necessarily the barycentre distribution.
Instead, since the modified or projected data will be used on a pre-trained classification or prediction model, the distribution of training data used to learn such a pre-trained model can be our target distribution, to preserve the classification or prediction performance.

\paragraph{Paper structure}  
The paper is organised as follows: 
Section~1.1 gives the motivating example of school admission. Section~1.2 introduces related work.
Section~2 presents the state-of-the-art in entropic regularised OT. 
Section~3 introduces our ``total repair'' and ``partial repair'' schemes with our algorithm. Sections~3.1-3.4 focus on a one-dimensional case (that is, only one unprotected attribute is considered) and Section~3.5 explains how to implement our schemes in higher-dimensional cases. 
Section~3.6 explains the choices of target distribution.
Sections~4-5 give the numerical results on synthetic data and real-world data, as well as a comparison with baselines.
The proofs of all lemmas and theorems can be found in Appendices.


\subsection{Motivating example}
Let us consider an example of school admission to illustrate the bias repair schemes in our framework as well as \cite{gordaliza2019obtaining,feldman2015certifying}.

\textcolor{black}{
Schools commonly use exam scores to make admission decisions based on a cut-off point: applicants with scores above this threshold are admitted. Suppose a school has employed this admissions algorithm for a long time, using a cut-off point determined by a large pool of applicants' scores. However, recent changes in the exam format or content have rendered the old data distribution out-dated.
When the school attempts to establish a new cut-off point, they face a scarcity of data due to the recent changes, resulting in only a few new scores available. Additionally, the change in exam format has temporarily suppressed performance, particularly affecting applicants from underprivileged groups who have fewer resources to adapt. Consequently, if the school continues to use the old cut-off point, they may inadvertently admit more applicants from privileged groups than usual.
To address this issue, the school seeks to project the new scores of both groups onto the old distribution, but they encounter the challenge of unknown group membership. Meanwhile, they cannot apply different cut-off points for different groups without knowing which applicants belong to which group.}

\begin{figure}[!htp]
\centering
\includegraphics[width=0.65\textwidth]{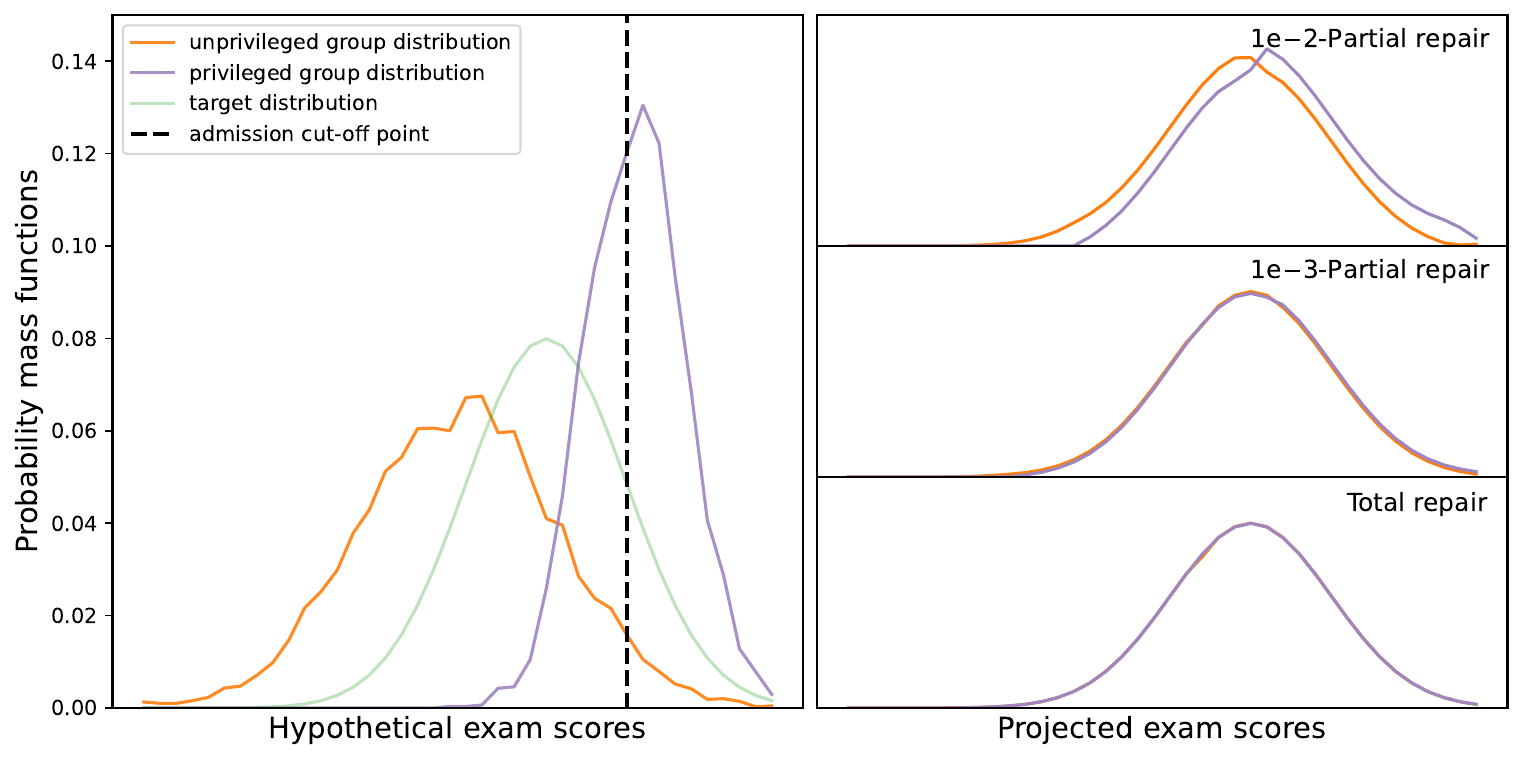}
\caption{\textbf{Left:} A cut-off point (vertical dark line) for an exam score is used to make school-admission decisions. With the hypothetical exam-score distributions of the unprivileged group (orange) and the privileged group (purple), this group-blind cut-off point would result in outcomes biased against the unprivileged group. 
If the score distributions of unprivileged group and the privileged group are projected closer to an target distribution (green), the same cut-off point can achieve equalised admission rates.
\textbf{Right:} The partial repair (top two) and total repair (bottom) schemes proposed in this paper, move the score distributions of unprivileged group (orange) and the privileged group (purple) closer to the pre-defined target distribution, without access to the protected attribute of each sample in training process.}
\label{fig:motivation}
\end{figure}

In Figure~\ref{fig:motivation}, the old cut-off point is shown with the vertical dashed line. 
There are two groups of applicants, divided by a protected attribute, e.g., gender, race. 
The left subplot of Figure~\ref{fig:motivation} shows that the score distribution of the privileged group (denoted by a purple curve) concentrates somewhere above the cut-off point.
On the contrary, only a small portion of the unprivileged group (denoted by an orange curve) passes the cut-off point.
If the cut-off point is used in a straightforward manner, the admission results are strongly biased against the unprivileged group, although the admission-decision algorithm does not take into account the protected attribute.
Then, instead of training another fairness-aware admission-decision model, the bias repair schemes will project the score distributions of both groups into an  target distribution (the green curve), for example, the old score distribution, such that the scores of both groups follow the same distribution that was used to train the old cut-off point. The target distribution is a ``median'' distribution in \cite{feldman2015certifying}, and a barycentre distribution in \cite{gordaliza2019obtaining}, and are not necessarily the same as the green curve.

The right subplots of Figure~\ref{fig:motivation} illustrate the effects of bias-repair schemes proposed in this paper, with the top two for ``partial repair'', and the bottom one for ``total repair''.
Now, the old admission cut-off point used on the projected scores will achieve equalised admission rates between the privileged and the unprivileged group.
Our extension eliminates the need for the protected attribute of each sample when computing the group-blind projection map and projecting both groups via this map.

\subsection{Related work}
\label{sec:related-work}

\paragraph{Fairness in machine learning}
There are various definitions and metrics of fairness, nicely surveyed in \cite{castelnovo2022clarification,pessach2022review,mehrabi2021survey}. Different fairness criteria may conflict with each other \cite{kim2020fact,binns2020apparent}.
The category of group fairness stands out as the most extensively studied, comprising demographic parity addressing disparate impact, fairness through unawareness \citep{fabris2023measuring} addressing disparate treatment, equal opportunity \citep{hardt2016equality} and predictive equality  \citep{zeng2022fair,chouldechova2017fair}, both of which address disparate mistreatment \citep{zafar2019fairness,zafar2017fairness}.
Fairness measures have been extended to individual fairness \citep{petersen2021post,dwork2012fairness}, procedural fairness \citep{grgic2018beyond}, as well as (path-specific) counterfactual fairness \citep{kusner2017counterfactual,chiappa2019path,nabi2022optimal,plecko2022causal,nilforoshan2022causal}.

Multiple studies have extensively examined the trade-offs that arise between fairness and accuracy in \cite{liu2022accuracy,mandal2020ensuring}, the trade-offs between group fairness and individual fairness in \cite{binns2020apparent}, and the trade-offs between fairness and privacy in \cite{chang2021privacy,cummings2019compatibility}.
There have been designs of systems to balance multiple potentially conflicting fairness measures \citep{kim2020fact,lohia2019bias}. 
These fairness criteria have been used in fair classification  \citep{zafar2019fairness,barocas-hardt-narayanan}, fair prediction \citep{plecko2022causal,chouldechova2017fair}, fair ranking \citep{narasimhan2020pairwise,feldman2015certifying}, fair policy \citep{plecko2023causal,chzhen2021unified,nabi2019learning}, and fair representation in \citep{zhao2022inherent,creager2019flexibly,zemel2013learning}.
In the long term, maintaining machine learning models adaptable to dynamic fairness concepts, due to evolving data distributions, or changing societal norms, has been considered  \citep{bechavod2023individually,jabbari2017fairness}.

\paragraph{Fairness without demographics}

\textcolor{black}{As noted, there is often a gap between unavailable demographic information in practice and what fairness methods assume is accessible. Most practitioners regard limited access to such information as a major barrier to implementing fairness techniques \citep{andrus2021we}.}
The challenge of unavailable demographic information has been partially relaxed to considering protected attributes only at training time \citep{oneto2020fairness,jiang2020wasserstein,quadrianto2017recycling,zafar2017fairnessBeyong}. This approach trains a model that is blind to protected attributes by employing regularisation during training to penalise performance discrepancies across different groups.
This line of research has recently been extended to require protected attributes only in a validation set \citep{elzayn2023estimating, chai2022self} to guide the training process. In the two-stage framework of \cite{liu2021just}, protected attributes in the validation set are employed to compute the worst-group validation error, which is then utilised for hyper-parameter tuning.

\color{black}
As introduced earlier in this section, the standard definitions of group fairness based on specific protected attributes, have been extensively recognised, researched, and are readily accessible in open-source toolkits.
In practical scenarios, certain approaches aim to ensure fairness towards unspecified minority groups, often using proxies, or some noisy estimates.
\cite{amini2019uncovering,oneto2019taking} introduce the latent variables as the proxies of unknown protected attributes.
Additionally, \cite{sohoni2020no} explore clustering for a similar purpose. 
\cite{zhao2022towards} minimises the correlation between proxies and model predictions to achieve fair classifier learning. 
Some methods employ distributional robust optimisation techniques to minimise the worst-case expected loss for groups identified by ambiguous protected attributes \citep{wang2020robust,sohoni2020no}, or by regions where the model incurs (computationally-identifiable) errors \citep{lahoti2020fairness,veldanda2023hyper}.


Further along the road, some fairness definitions that does not focus on groups may better suit specific contexts, while necessitating more specialised expertise. 
``Fairness through awareness'' could be one example that requires context-specific definitions of similarity among individuals.
\cite{liu2023group} avoids any form of group memberships and uses only pairwise similarities between individuals to define inequality in outcomes, using the common property of social networks that individuals sharing similar attributes are more likely to be connected to each other than individuals that are dissimilar.
Additionally, a survey in ``Fairness Without Demographic Data'' \citep{ashurst2023fairness} also discusses trusted third parties and cryptographic solutions to encourage the collection of protected attributes. These methods, e.g., \cite{veale2017fairer}, focus on protecting protected attributes from misuse or use without informed consent, while also offering guidance or assessment for AI developers.
\color{black}

\color{black}
\paragraph{Domain Adaptation}
The concept of projecting data distributions to a target is a well-researched area within domain adaptation, a subfield of transfer learning that utilises data from one or more relevant source domains to perform new tasks in a target domain. Various factors, such as different poses in object recognition or varying exam content in school admissions, often lead to distribution changes or domain shifts between source and target domains.
According to \cite{pan2010survey}, domain shifts can be categorised into two types: domain divergence, which refers to distribution shifts or differences in feature space, and task divergence, which concerns variations in the conditional distribution of the target over features or label space. In homogeneous settings of domain adaptation, it is assumed that the tasks remain the same, with only a distribution shift occurring between the source and target domains, which aligns closely with the scenario presented in this paper.

Numerous methods have been proposed to tackle domain shifts in domain adaptation, and these can be broadly divided into two categories, as noted by \cite{WANG2018deep}. Instance-based domain adaptation involves re-weighting the source samples and training on these weighted samples. In contrast, feature-based domain adaptation focuses on reducing the distance between domains in a latent space. This approach includes techniques such as feature augmentation \citep{hal2011domain} and transfer component analysis \citep{pan2011domain}, which identify a nonlinear feature map within a reproducing kernel Hilbert space by employing maximum mean discrepancy. In the subspace defined by these transfer components, the properties of the data are preserved, and the distributions across different domains become more closely aligned. Consequently, with these new representations, standard machine learning methods can be effectively applied to train classifiers or regression models in the source domain for subsequent use in the target domain.
In the deep domain adaptation community, various methods that employ deep networks include those that fine-tune the network model using discrepancy-based regularisations \citep{long2015learning}, as well as techniques that reconstruct source or target samples through encoders for representation learning and decoders for data reconstruction \citep{bousmalis2016domain}.
\color{black}

\paragraph{Optimal transport} 
We refer to \cite{peyre2019computational} for a comprehensive overview of the history of optimal transport (OT).
It is a method to find a coupling that moves mass from one distribution to another at the smallest possible cost, known as the Monge problem, relaxed into so-called Kantorovich relaxation. 
The relaxation, with an entropic regularisation to promote smoothness of the map, leads to iterations of simple matrix-vector products of the Sinkhorn--Knopp algorithm \citep{sinkhorn1967concerning}. 
Often, the equality constraints in entropy-regularised OT are softened by minimising the dual norm of the differences between both sides of equality constraints; see  \cite{chapel2021unbalanced,pham2020unbalanced,chizat2018scaling} for unbalanced OT and \cite{chizat2018unbalanced} for dynamic unbalanced OT. 
Pioneered by \cite{gangbo1998optimal,agueh2011barycenters},
another direction is to explore barycentres of Wasserstein distance of multiple distributions with respect to some weights, which is a representation of the mean of the set of distributions, with several follow-up works in \cite{nath2020statistical,dard2016relevance,seguy2015principal}.
On the computational side, the simplest algorithm that exploits Kullback-Leibler (KL) divergence is the iterative Bregman projections \citep{bregman1967relaxation}, which is related to the Sinkhorn algorithm. 
The generic OT considering finitely many arbitrary convex constraints can be solved by Dykstra's algorithm with Bregman projections \citep{bauschke2000dykstras,benamou2015iterative}, or generalised scaling algorithm \citep{chizat2018scaling}. The mini-batch OT can be used to ease the computational burden  \citep{sommerfeld2019optimal}.
The classic OT formulation with entropic regularisation penalty, together with its generalisations is explained in Section~\ref{sec:regularised-OT}.

OT has been widely used in domain adaptation \citep{pmlr-v180-turrisi22a,pmlr-v162-nguyen22e,courty2017joint} by computing a map that project the distributions of multiple datasets into a common representation. See \cite{montesuma2023recent} for a survey.
The same principle of using OT as a projection map could be used to mitigate algorithmic bias in machine-learning models  \citep{gordaliza2019obtaining,silvia2020general,yang2022obtaining}. 
Utilisation of OT as a loss or a metric has emerged as a notable  trending subject within bias mitigation \citep{buyl2022optimal,risser2022tackling,jiang2020wasserstein,oneto2020fairness} and bias detection \citep{si2021testing,black2020fliptest}.



\section{Background and Notation}
\label{sec:background}
Let $N$ be the number of discretisation points, and $P\in\bb{R}^N$ be vectors in the probability simplex:
\begin{equation}
\Sigma_N:=\left\{P\in\bb{R}^N_+\Bigg|\sum^N_{i=1} P_i=1\right\}.
\end{equation}
The set of all admissible or feasible couplings between $(P,Q)\in\Sigma^2_N$ is defined as 
\begin{equation}
\Pi(P,Q):=\{\gamma\in\bb{R}^{N\times N}_+\mid \gamma\mathbb{1}=P,\gamma^{\tr}\mathbb{1}=Q\},
\label{equ:Pi-define}
\end{equation}
where $\mathbb{1}$ is the $N$-dimensional column vector of ones, $\gamma^{\tr}$ is the transpose of $\gamma$. 
Further, we define the entropy of a coupling $\gamma$ as:
\begin{equation}
E(\gamma):=-\sum_{i,j=1}^N \gamma_{i,j}(\log(\gamma_{i,j})-1), \label{equ:entropy-define}
\end{equation}
which is concave function and we use the convention $0\log 0=0$. The Kullback-Leibler (KL) divergence between the coupling $\gamma\in\bb{R}^{N\times N}_+$ and a positive reference matrix $\xi\in\bb{R}^{N\times N}_{++}$ (i.e., $\xi_{i,j}>0$) is defined as:
\begin{equation}
\kl{\gamma}{\xi}:=\sum_{i,j=1}^N \gamma_{i,j}\left(\log\left(\frac{\gamma_{i,j}}{\xi_{i,j}}\right) -1 \right).\label{equ:kl-define}
\end{equation}
Given a convex set $\mathcal{C}\in\bb{R}^{N\times N}$, the KL projection is 
\begin{equation}
\prox^{KL}_{\mathcal{C}}(\xi):=\arg\min_{\gamma\in\mathcal{C}} \kl{\gamma}{\xi},\label{equ:kl-projec-define}
\end{equation}
which is uniquely defined since KL divergence is a strictly convex and coercive function and, $\mathcal{C}$ is a convex set. See Lemma~\ref{lem:subgradient-optimality} for more details. 

\subsection{Regularised Optimal Transport}
\label{sec:regularised-OT}
Referring to \cite{benamou2015iterative}, we introduce the state of the art in OT.
The discrete entropic regularisation OT problem has the form:
\begin{equation}
W_{\epsilon}(P,Q):=\min_{\gamma\in\Pi(P,Q)} \langle C,\gamma\rangle -\epsilon E(\gamma),
\label{equ:entropic-reg-OT-define}
\end{equation}
where $C\in\bb{R}^{N\times N}$ is the cost matrix, $\epsilon>0$ is the entropic regularisation parameter and $\langle\cdot,\cdot\rangle$ is the inner product. This formulation is equivalent to
\begin{equation}
\begin{split}
W_{\epsilon}(P,Q)&=\min_{\gamma\in\Pi(P,Q)} \sum^N_{i,j=1} C_{i,j}\gamma_{i,j}+\epsilon\gamma_{i,j}\left(\log (\gamma_{i,j}) -1 \right)\\
&=\min_{\gamma\in\Pi(P,Q)}\epsilon\sum^N_{i,j=1} \gamma_{i,j}\left(\log(\gamma_{i,j})-\log(\exp(-C_{i,j}/\epsilon)) -1 \right)\\
&=\min_{\gamma\in\Pi(P,Q)} \epsilon\;\kl{\gamma}{\xi}, \quad\textrm{where}\quad\xi=\exp{(-C/\epsilon)}.
\end{split} 
\label{equ:KL-equivalent}
\end{equation}

Further, the set of admissible (feasible) couplings in Equation~\eqref{equ:Pi-define} could be formulated as the intersection of two affine subspaces: \textcolor{black}{$\Pi(P,Q)=\mathcal{C}_1\cap\mathcal{C}_2\neq\emptyset$}, where
\begin{equation}
\mathcal{C}_1:=\{\gamma\in\bb{R}^{N\times N}_+\mid\gamma\mathbb{1}=P\},\quad\mathcal{C}_2:=\{\gamma\in\bb{R}^{N\times N}_+\mid\gamma^{\tr}\mathbb{1}=Q\}.\label{equ:convex-set-1}
\end{equation}

Generally, when the convex sets $\mathcal{C}_{\ell},\ell\geq 1$ are affine subspaces,
initialise $\gamma^{(0)}:=\xi=\exp{(-C/\epsilon)}$, then conduct iterative KL projections: for $n>0$
\begin{equation}
\gamma^{(n+1)}=\prox^{KL}_{\mathcal{C}_n} (\gamma^{(n)}),\label{equ:iterative-Bregman}
\end{equation} 
where $\mathcal{C}_n\equiv\mathcal{C}_{1+(n\mod 2)}$. 
The iterations in Equation~\eqref{equ:iterative-Bregman} can converge to the unique solution of $W_{\epsilon}(P,Q)$, as $n\to+\infty$ (Fact~1.4 in \cite{bauschke2020dykstra}, \cite{benamou2015iterative,bregman1967relaxation}). Note that the KL projections used in the paper, is a special case of the general Bregman projections as in Definition~\ref{def:app-Bregman}.

\paragraph{Inequality constraints}
When the convex sets are not affine subspaces, iterative Bregman projections do not converge to $W_{\epsilon}(P,Q)$ (Fact~1.2 in \cite{bauschke2020dykstra}). Instead, Dykstra's algorithm is known to converge when used in conjunction with Bregman divergences \textcolor{black}{for arbitrary closed convex sets with a non-empty intersection} (cf. Appendix~\ref{app:Dykstra} or \cite{bauschke2020dykstra,bauschke2000dykstras}).
Random Dykstra algorithm converges linearly in expectation for the general feasible sets satisfying Slater's condition \citep{necoara2022linear}.

It could be used to solve OT problem with inequality constraints:
\begin{example}\label{exp:partial-OT}
The partial OT \citep{le2022multimarginal,caffarelli2010free,figalli2010optimal} only needs to transport a given fraction of mass and the two marginals $P,Q$ do not need to have the same total mass:
\begin{equation*}
\min_{\gamma\in\bb{R}^{N\times N}_+}\{\langle C,\gamma\rangle -\epsilon E(\gamma)\mid \gamma\mathbb{1}\leq P,\gamma^{\tr}\mathbb{1}\leq Q, \mathbb{1}^{\tr}\gamma\mathbb{1}=\eta\},
\end{equation*}
where the constant $\eta\in[0,\min\{P^{\tr}\mathbb{1},Q^{\tr}\mathbb{1}\}]$ denotes the fraction of mass needed to transport. Similarly, this problem can be reformulated as Equation~\eqref{equ:KL-equivalent} but with $\gamma$ falling into the intersection of three convex sets:
\begin{equation*}
\mathcal{C}_1:=\{\gamma\in\bb{R}^{N\times N}_+\mid\gamma\mathbb{1}\leq P\},\;\mathcal{C}_2:=\{\gamma\in\bb{R}^{N\times N}_+\mid\gamma^{\tr}\mathbb{1}\leq Q\},\;\mathcal{C}_3:=\{\gamma\in\bb{R}^{N\times N}_+\mid\mathbb{1}^{\tr}\gamma\mathbb{1}=\eta\}.
\end{equation*}
\end{example}

\begin{example}\label{exp:capacity-OT}
The capacity constrained OT, pioneered by \cite{korman2015optimal,korman2013insights}, is imposed by an extra upper bound on the amount of mass transported from $i$ to $j$. Its formulation reads:
\begin{equation*}
\min_{\gamma\in\bb{R}^{N\times N}_+}\{\langle C,\gamma\rangle -\epsilon E(\gamma)\mid \gamma\mathbb{1}= P,\gamma^{\tr}\mathbb{1}=Q, \gamma\leq\Theta \},
\end{equation*}
where $\Theta \in\bb{R}^{N\times N}_+$ denotes the upper bound. We 
equivalently write the feasible set into the intersection of the following three convex sets:
\begin{equation*}
\mathcal{C}_1:=\{\gamma\in\bb{R}^{N\times N}_+\mid\gamma\mathbb{1}=P\},\;\mathcal{C}_2:=\{\gamma\in\bb{R}^{N\times N}_+\mid\gamma^{\tr}\mathbb{1}=Q\},\;\mathcal{C}_3:=\{\gamma\in\bb{R}^{N\times N}_+\mid\gamma\leq\Theta \}.
\end{equation*}
\end{example}

\paragraph{Computation of KL projections}
In general, it is impossible to directly compute the KL projections in closed form, so some form of subiterations are required. Notably, the convex set with the form in Equation~\eqref{equ:convex-set-1}, Example~\ref{exp:partial-OT}-\ref{exp:capacity-OT} can be computed by some matrix-vector multiplications \citep{benamou2015iterative}, with the $\min$, division operators being element-wise:
\begin{equation*}
\prox^{KL}_{\mathcal{C}}(\bar{\gamma})=
\begin{cases}
\diag{\frac{P}{\gamma\mathbb{1}}}\bar{\gamma} & \textrm{if } \mathcal{C}=\{\gamma\in\bb{R}^{N\times N}_+\mid\gamma\mathbb{1}=P\},\\
\bar{\gamma}\diag{\frac{Q}{\bar{\gamma}^{\tr}\mathbb{1}}} & \textrm{if } \mathcal{C}=\{\gamma\in\bb{R}^{N\times N}_+\mid\gamma^{\tr}\mathbb{1}=Q\},\\
\diag{\min\left\{\mathbb{1},\frac{P}{\bar{\gamma}\mathbb{1}}\right\}}\bar{\gamma} & \textrm{if } \mathcal{C}=\{\gamma\in\bb{R}^{N\times N}_+\mid\gamma\mathbb{1}\leq P\},\\
\bar{\gamma}\diag{\min\left\{\mathbb{1},\frac{Q}{\bar{\gamma}^{\tr}\mathbb{1}}\right\}} & \textrm{if } \mathcal{C}=\{\gamma\in\bb{R}^{N\times N}_+\mid\gamma^{\tr}\mathbb{1}\leq Q\},\\
\bar{\gamma}\frac{\eta}{\mathbb{1}^{\tr}\bar{\gamma}\mathbb{1}} & \textrm{if } \mathcal{C}=\{\gamma\in\bb{R}^{N\times N}_+\mid\mathbb{1}^{\tr}\gamma\mathbb{1}=\eta\},\\
\min\{\bar{\gamma},\Theta \} & \textrm{if } \mathcal{C}=\{\gamma\in\bb{R}^{N\times N}_+\mid\gamma\leq\Theta \},
\end{cases}
\end{equation*}
where the proof for the first two cases can be found in Appendix~\ref{app:lem:c_1&c_2}.

\section{Our Framework}

Recall the school admission example introduced in Section~1.1. 
The school receive the exam score ($X$), but not the protected attribute ($S$) of each applicant. So, the source data would be all applicants' exam scores. Due to $S$ being not observed, the school wants to find a $S$-blind projection map ($\mathcal{T}$), to project these scores to the target data, where the two groups of applicants have the same (``total repair'') or equalised (``partial repair'') distributions of scores. 
In other words, the school wants to achieve equalised odds of admission between the privileged and unprivileged groups in the target data, using the pre-trained admission-decision policy.

\subsection{Definitions}

Let us introduce our ``total repair'' and ``partial repair'' schemes, starting with defining the protected attribute ($S$), distributions of the exam scores ($X$) and $S$-blind projection map ($\mathcal{T}$):

\begin{definition}[A protected attribute]
Let $S\in\bb{N}$ be an integer random variables of a  protected attribute (e.g., race, gender). Let $P^{S}_s:=\p{S=s}$ be the probability of $S=s$. Its support is defined as 
\begin{equation}
\supp{S}:=\{s\in\bb{N}\mid \p{S=s}>0\}.
\end{equation}
\end{definition}

\begin{definition}[Source variables of an unprotected attribute]
Let $X,X_{s},s\in\supp{S}$ be scalar discrete random variables of an unprotected attribute (e.g., income, credit scores), with their supports $\supp{X_s}\subseteq\supp{X}\subset\bb{R}$, for $s\in\supp{S}$.
Let $\supp{X}$ be $N$ discretisation points.
Their probability distributions are taken from the probability simplex $P^{X},P^{X_s}\in\Sigma_N$:
\begin{equation}
P^{X}_i:=\p{X=i},\quad P^{X_s}_i:=\pover{X=i}{S=s},
\end{equation}
where $\p{X=i}$ is the probability of $X=i$ and $\pover{X=i}{S=s}$ is the conditional probability of $X=i$ when $S=s$.
\end{definition}

To an exploration, throughout the paper we only consider the case of two groups, i.e., $\supp{S}=\{s_0,s_1\}$.
we assume $X$ includes one unprotected attribute till Section~3.4 and the extension to multiple unprotected attributes will be discussed in Section~3.5.

Next, we define the target distribution, for instance, the ideal score distribution that we want to achieve in the target data. 
The choices of target distribution will be mentioned in Section~3.6.

\begin{definition}[Target variables of the unprotected attribute]
Let $\td{X},\td{X}_{s},s\in\supp{S}$ be scalar discrete random variables, with their supports $\supp{\td{X}_s}\subseteq\supp{\td{X}}\subset\bb{R}$, for $s\in\supp{S}$.
Let $\supp{\td{X}}$ be $N$ discretisation points.
Their probability distributions are taken from the probability simplex $P^{\td{X}},P^{\td{X}_s}\in\Sigma_N$:
\begin{equation}
P^{\td{X}}_j:=\p{\td{X}=j},\quad P^{\td{X}_s}_j:=\pover{\td{X}=j}{S=s},
\end{equation}
\end{definition}


In our setting, we would like to map the source data, i.e., samples from the tuple $(X,S)$ to the target data, i.e., samples from the tuple $(\td{X},S)$, via a projection map $\mathcal{T}$, that is defined to be $S$-blind, and induced by a coupling $\gamma\in\Pi(P^{X},P^{\td{X}})$:

\begin{definition}[Projection]\label{def:projection}

Once the coupling $\gamma\in\Pi(P^{X},P^{\td{X}})$ has been computed, to find the map that transports source data to target data, we define the projection $\mathcal{T}$:
\begin{equation*}
\begin{split}
\mathcal{T}:\supp{X}&\to\supp{\td{X}}\times\bb{R}_{+}\\
i&\mapsto (j,w_{i,j}),\forall j\in\supp{\td{X}}
\end{split}
,\quad\textrm{ where } w_{i,j}=\frac{\p{X=i,\td{X}=j}}{\p{X=i}}=\frac{\gamma_{i,j}}{P^X_{i}}.
\end{equation*}
A sample $(i,s)$ is hence split into a sequence of weighted samples $\{(j,w_{i,j},s)\}_{j\in\supp{\td{X}}}$.
Very importantly, we stress that the projection do not change $s$ and any other attributes even though $s$ is not observed. 
\end{definition}

\begin{example}\label{exp:projection-1}
Given a sample $(i,s)$ and
$\gamma_{i,j}=P^{X}_i$. Thus $\gamma_{i,j'}=0$ for $j'\in\supp{\td{X}}\setminus\{j\}$ and 
\begin{equation*}
\mathcal{T}(i)=(j,1).
\end{equation*}
Hence, this sample is transported to $(j,1,s)$. 
\end{example}
\begin{example}\label{exp:projection-2}
Given a sample $(i,s)$ and
$\gamma_{i,j}=\gamma_{i,j'}=P^{X}_i/2$.
Thus $\gamma_{i,j^{\dag}}=0$ for $j^{\dag}\in\supp{\td{X}}\setminus\{j,j'\}$ and
\begin{equation*}
\mathcal{T}(i)=
\begin{array}{c}
(j,1/2)\\
(j',1/2)
\end{array}.
\end{equation*}
Hence, this sample is split into $(j,1/2,s)$ and $(j',1/2,s)$.
\end{example}

\paragraph{Observations} 
So far, we have defined the projection map $\mathcal{T}$ in the formal way. We first observe something interesting:

Let $\supp{S}=\{s_0,s_1\},\supp{X}=\{i,i'\}$, $\supp{\td{X}}=\{j,j'\}$. Then, our data only include six samples: 
\begin{align*}
\textcolor{orange}{(i,s_0)},\textcolor{violet}{(i,s_1),(i,s_1)},\\
\textcolor{orange}{(i',s_0),(i',s_0)},\textcolor{violet}{(i',s_1)}.
\end{align*}
From the source data, we can directly compute
$P^{X}=[1/2,1/2]^{tr}$, $P^{X_{s_0}}=[1/3,2/3]^{tr}$ and $P^{X_{s_1}}=[2/3,1/3]^{tr}$.

If we set entries $\gamma_{i,j}=\gamma_{i',j'}=1/2$, and $\gamma_{i,j'}=\gamma_{i',j}=0$, following Example~\ref{exp:projection-1}, the projected data become:
\begin{align*}
\textcolor{orange}{(j,1,s_0)},\textcolor{violet}{(j,1,s_1),(j,1,s_1)},\\
\textcolor{orange}{(j',1,s_0),(j',1,s_0)},\textcolor{violet}{(j',1,s_1)},
\end{align*}
such that $P^{\td{X}}=P^{X}$, $P^{\td{X}_{s_0}}=P^{X_{s_0}}$, $P^{\td{X}_{s_1}}=P^{X_{s_1}}$. Nothing has changed.
However, if we set entries $\gamma_{i,j}=\gamma_{i',j'}=\gamma_{i,j'}=\gamma_{i',j}=1/4$, following Example~\ref{exp:projection-2}, the projected data become:
\begin{align*}
\textcolor{orange}{(j,1/2,s_0),(j,1/2,s_0),(j,1/2,s_0)},\textcolor{violet}{(j,1/2,s_1),(j,1/2,s_1),(j,1/2,s_1)},\\
\textcolor{orange}{(j',1/2,s_0),(j',1/2,s_0),(j',1/2,s_0)},\textcolor{violet}{(j',1/2,s_1),(j',1/2,s_1),(j',1/2,s_1)},
\end{align*}
such that $P^{\td{X}}=P^{X}=P^{\td{X}_{s_0}}=P^{\td{X}_{s_1}}$.
This is a trivial example indeed, but it hints at the possibility of manipulating $P^{\td{X}_{s_0}}$ and $P^{\td{X}_{s_1}}$ by a well-designed $S$-blind map.

Next, we explain how to extend the observations to nontrivial cases.

\subsection{Total Repair}

In the school admission example, even when the protected attribute $S$ is not observed, and the map is $S$-blind, we would like to achieve some parity between groups partitioned by $S$, in the exam score distributions.
Formally,


\begin{definition}[Total repair]\label{def:total-repair}
Following the definition of total repair in \cite{gordaliza2019obtaining}, 
we say that total repair is satisfied if 
\begin{equation}
P^{\td{X}_{s}}=P^{\td{X}_{s'}},\forall s,s'\in\supp{S}.
\end{equation}
Note that when total repair is satisfied, it holds $P^{\td{X}_{s}}=P^{\td{X}_{s'}}=P^{\td{X}}$.
\end{definition}
\begin{lemma}\label{lem:tx|us}
If the coupling $\gamma\in\Pi(P^{X},P^{\td{X}})$ is given, 
\begin{equation}
P^{\td{X_s}}=\gamma^{\tr}\frac{P^{X_s}}{P^{X}}, \quad P^{\td{X}}=\gamma^{\tr}\mathbb{1},
\end{equation}
where the division operator is element-wise.
\begin{proof}
See Appendix~\ref{app:lem:tx|us}.
\end{proof}
\end{lemma}

\paragraph{Intuitive verification}
Alternatively, we can think about the source data where the probability equals to the proportion.
We select all samples that belong to group $s$: $\{(i_m,s)\}_{m=1,\dots,M}$. Since the projection does not change $s$, a sample $(j,s)$ in the target data only originates from sample $(i,s),i\in\supp{X}$ in the source data.
Now, we equivalently group the same samples and use the number of the same samples as the weight. For instance, if we have five samples of $(i,s)$, we only use one weighted sample $(i,w_i=5,s)$ to represent all of them. Since we assume that probability equals portion, the source data can be rewritten as
\begin{equation}
\{(i,w_i,s)\}_{i\in\supp{X_s}},\label{equ:weighted-source-data}
\end{equation}
where $w_i=M\times\pover{X=i}{S=s}$ for $i\in\supp{X_s}$. From the weighted source data, we can compute $\pover{X=i}{S=s}=w_i/M$. 
Recall Definition~\ref{def:projection}: a sample $(i,s)$, or equivalently a weighted sample $(i,w_i=1,s)$, is transported to a sequence of weighted samples $\{(j,1\times w_{i,j},s)\}_{j\in\supp{\td{X}}}$. 
Then the weighted sample $(i,w_i,s)$ is transported to $(j,w_i\times w_{i,j},s)$ for $j\in\supp{\td{X}}$. After projecting all samples in weighted source data (Equation~\eqref{equ:weighted-source-data}), the projected data become
\begin{equation*}
\{(j,w_i\times w_{i,j},s)\}_{i\in\supp{X_s}, j\in\supp{\td{X}}}.
\end{equation*} 
Hence, we can compute
\begin{equation*}
\pover{\td{X}=j}{S=s}=\sum_{i\in\supp{X_s}} w_{i,j}\times w_{i}/M=\sum_{i\in\supp{X_s}}\pover{X=i}{S=s}\frac{\p{X=i,\td{X}=j}}{\p{X=i}}.
\end{equation*}
Rewrite the above equation in matrix form, and setting the undefined conditional probability to $\pover{X=i}{S=s}=0$ for $i\in\supp{X}\setminus\supp{X_s}$, we get the same results as in Lemma~\ref{lem:tx|us}.

\color{black}
Next, we explore what kind of couplings can guarantee total repair in the target data, such that the total repair property is transformed into a constraint imposed on the coupling.
In the case of binary  protected attributes, i.e., $\supp{S}=\{s_0,s_1\}$, we
define the vector $V\in\bb{R}^{N}$ as
\begin{equation}
V:=\frac{P^{X_{s_0}}-P^{X_{s_1}}}{P^{X}},\label{equ:V-def}
\end{equation}
which is the most important input in our framework, that only needs $P^{X_{s_0}},P^{X_{s_1}}$ and $P^{X}$ to be computed. Note that $P^{X}$ is already known due to observability of the attribute $X$ in source data.
While $S$ is not observed, $P^{X_{s_0}},P^{X_{s_1}}$ can be obtained from the population-level information regarding the distributions of attributes for both groups ($s_0$ and $s_1$), given the crucial assumption that the source data represents unbiased samples from the broader population. This assumption of unbiased sampling ensures that statistical properties of the population carry over.
Then, if one wishes to achieve total repair, the coupling should satisfy the following:
\begin{theorem}[Total repair to a constraint on the coupling $\gamma$]
\label{pro:binary_condition}
In the case of binary  protected attributes, i.e., $\supp{S}=\{s_0,s_1\}$, it holds that
\begin{equation}
P^{\td{X_0}}-P^{\td{X_1}}=\gamma^{\tr} V=\mathbb{0},
\label{equ:difference2rv}
\end{equation}
where the division, subtraction operations are element-wise, and $\mathbb{0}\in\bb{R}^{N\times 1}$ is a zero vector.
\begin{proof}
See Appendix~\ref{app:pro:binary_condition}.
\end{proof}
\end{theorem}
\color{black}

There are some interesting properties of the vector $V$ that will be useful for building important lemmas in the following sections.
\begin{remark}[Properties of the vector $V$]\label{rem:v-property}
The vector $V$, defined in Equation~\eqref{equ:V-def}, has the following properties.\begin{enumerate}[(i)]
\item $(P^X)^{\tr} V=0$.
\item \textcolor{black}{Non-zero entries of the vector $V$ are neither all negative nor all positive.}
\item For a vector $P\in\bb{R}^N$, let the $l_1$ norm of this vector be $\|P\|_1:=\sum_{i=1}^N |P_i|$.
If the norm $\|\frac{1}{P^X}\|_1$ is finite, then the norm $\|V\|_1$ is finite, with division being element-wise.
\end{enumerate}
\begin{proof}
Property (i) comes from the definition: $(P^X)^{\tr} V=\sum_{i\in\supp{X}}P^{X_{s_0}}_i-P^{X_{s_1}}_i=0$. See Appendix~\ref{app:rem:v-property} for details.
\end{proof}
\end{remark}


\subsection{Partial Repair}

The projection in Definition~\ref{def:projection} consists of changing the attributes and weights of samples in source data, while practical implementation may necessitate maintaining data distortion levels, i.e., the inner product $\langle C,\gamma\rangle$, within predefined acceptable thresholds.
From a practical point of view, we consider partial repair, as relaxations of the constraint in Theorem~\ref{pro:binary_condition}.

\begin{assumption}[Finite $l_1$ norm]\label{ass:finite}
The norm $\|\frac{1}{P^X}\|_1$ is finite, where the division operator is element-wise.
\end{assumption}

\color{black}
By Assumption~\ref{ass:finite}
and property (iii) in Remark~\ref{rem:v-property}, the vector $V$ has a finite $l_1$-norm without infinite entries.
Hence, we can find a non-negative vector $\Theta\in\bb{R}^N_+$ such that
\begin{equation}
\Theta_j\leq\sum_{i\in\supp{X}} \gamma_{i,j}V_i\leq \Theta_j,\quad \forall j\in\supp{\td{X}}.\label{equ:Theta-def}
\end{equation}

We first define partial repair before justifying it. 
\begin{definition}[Partial repair]\label{def:partialrepair}
Given a non-negative vector $\Theta\in\bb{R}^{N}_+$,
the $\Theta$-repair is satisfied if Equation~\eqref{equ:Theta-def} holds. 
\end{definition}
\color{black}
Note that $\mathbb{0}$-repair is equivalent to total repair.

To measure the violation of the total repair property, we introduce the TV distance and analyse it between $P^{\td{X}_{s_0}}$ and $P^{\td{X}_{s_1}}$ within the context of partial repair.

\begin{definition}[Total variation distance (abbr. TV distance)]\label{def:tvdistance}
Given two discrete probability distributions $P$, $Q$ over $\supp{\td{X}}$, the TV distance $\tv{P}{Q}$ between $P$ and $Q$ is defined as:
\begin{align}
\tv{P}{Q}:=\frac{1}{2}\sum_{j\in\supp{\td{X}}}|P_j-Q_j|=\frac{1}{2}\|P-Q\|_1.
\end{align}
\end{definition}

\color{black}
\begin{lemma}\label{lem:S-wise TV} 
When $\Theta$-repair in Definition~\ref{def:partialrepair} is satisfied, the TV distance between $P^{\td{X}_{s_0}}$ and $P^{\td{X}_{s_1}}$ is bounded by
\begin{equation}
\tv{P^{\td{X}_{s_0}}}{P^{\td{X}_{s_1}}}=\frac{\left\|\gamma^{\tr}V\right\|_1}{2}\leq \frac{\|\Theta\|_1}{2}.\end{equation}
\begin{proof}
Theorem~\ref{pro:binary_condition} shows that $P^{\td{X_0}}-P^{\td{X_1}}=\gamma^{\tr} V$. 
See Appendix~\ref{app:lem:S-wise TV} for details.
\end{proof}
\end{lemma}
\color{black}
Note that computing the value of the upper bound for TV distance needs to consider the dimension of $\Theta$, i.e, $N$.


\subsection{Formulations and algorithms}
Lemma~\ref{lem:S-wise TV} implies that
in the standard formulation of regularised OT, if we were able to add the extra constraint in Equation~\eqref{equ:Theta-def}, this optimal solution $\gamma^*$ would achieve $\Theta$-repair or total repair if $\Theta=\mathbb{0}$.
The revised formulation reads:
\begin{align}
\inf_{\gamma\in\Pi(P^{X},P^{\td{X}})} \left\{\langle C,\gamma\rangle-\epsilon E(\gamma)\Bigg|-\Theta_j\leq \sum_{i\in\supp{X}} \gamma_{i,j}V_i\leq\Theta_j,\forall j\in\supp{\td{X}}\right\},\label{equ:our-formulation-1}
\end{align}
where $\epsilon>0$ and $\Theta\in\bb{R}^{N}_+$.

\color{black}
Given that the additional constraints in this constrained OT, as described in Equation~\eqref{equ:our-formulation-1}, are convex, we can apply Dykstra's algorithm, introduced in Section~\ref{sec:background}, if the feasible region is non-empty.
We define the set of all admissible couplings in Equation~\eqref{equ:our-formulation-1}, i.e., the feasible region:
\begin{equation}
\Pi_{\Theta }(P^{X},P^{\td{X}}):=\{\gamma\in\Sigma^2_N\mid \gamma\mathbb{1}=P^{X}, \gamma^{\tr}\mathbb{1}=P^{\td{X}},\left|\gamma^{\tr} V\right|_j\leq\Theta_j,\forall j\in\supp{\td{X}}\},
\end{equation}
where $\left|\gamma^{\tr} V\right|_j:=\left|\sum_{i\in\supp{X}}\gamma_{i,j}V_i\right|$.
 We now proceed to verify the non-emptiness.

\begin{lemma}[The feasible set is non-empty]\label{lem:existence}
Given two non-negative vectors $\Theta',\Theta\in\mathbb{R}^{N}_+$, with $\Theta'_j\leq\Theta_j$ for all $j\in\supp{\td{X}}$, the following sequence holds
\begin{equation}
\emptyset\neq\Pi_{\mathbb{0}}(P^{X},P^{\td{X}})\subseteq\Pi_{\Theta'}(P^{X},P^{\td{X}})\subseteq\Pi_{\Theta}(P^{X},P^{\td{X}})\subseteq\Pi(P^{X},P^{\td{X}}).
\end{equation}
\begin{proof}
Remark~2.13 in \cite{peyre2019computational} states that $\Pi(P^{X},P^{\td{X}})$ is non-empty because the coupling $P^{X}\otimes P^{\td{X}}\in\Pi(P^{X},P^{\td{X}})$, where $\otimes$ is the outer product. Using Property (i) in Remark~\ref{rem:v-property}, we show that this coupling $P^{X}\otimes P^{\td{X}}\in\Pi_{\mathbb{0}}(P^{X},P^{\td{X}})$. See Appendix~\ref{app:lem:existence}.
\end{proof}
\end{lemma}
Therefore, the feasible region of Equation~\eqref{equ:our-formulation-1} for arbitrary  $\Theta\in\mathbb{R}^{N}_+$ is non-empty.
\color{black}
The minimum exists due to this and the coercivity of the objective function. 
Following the same reasoning in Equation~\eqref{equ:KL-equivalent}, we equivalently rewrite Equation~\eqref{equ:our-formulation-1} into 
\begin{align}
\min_{\gamma\in\Pi_{\Theta }(P^{X},P^{\td{X}})} \kl{\gamma}{\xi}, \textrm{ where }\xi=\exp{(-C/\epsilon)},
\label{equ:our-formulation-2}
\end{align}
and $\Pi_{\Theta }(P^{X},P^{\td{X}})=\bigcap^{3}_{\ell=1}\mathcal{C}_{\ell}$ is the intersection of three convex sets:
\begin{equation}
\begin{split}
\mathcal{C}_1=\{\gamma\in\bb{R}^{N\times N}_+\mid\gamma\mathbb{1}=P^{X}\},&\mathcal{C}_2=\{\gamma\in\bb{R}^{N\times N}_+\mid\gamma^{\tr}\mathbb{1}=P^{\td{X}}\},\\
\mathcal{C}_3=\{\gamma\in\bb{R}^{N\times N}_+&\mid -\Theta\leq\gamma^{\tr}V\leq\Theta\}.
\end{split}
\label{equ:convex-sets}
\end{equation}
Note that $\mathcal{C}_3$ is equivalent to $\{\gamma\in\bb{R}^{N\times N}_+\mid \gamma^{\tr}V=\mathbb{0}\}$ if $\Theta=\mathbb{0}$.

Then, applying Dykstra's algorithm, we propose Algorithm~\ref{alg:Dykstra} to solve Equations~(\ref{equ:our-formulation-2}-\ref{equ:convex-sets}).
The inputs consist of the support $\supp{X}$ (resp. $\supp{\td{X}}$) and probability distribution $P^{X}$ (resp. $P^{\td{X}}$) of variables $X$ (resp. $\td{X}$); the number of discretised point of these supports $N$; the vectors $V$, $\Theta$; the cost matrix $C$; entropic regularisation parameter $\epsilon>0$, the number of iterations $K$ and a very small number $\varepsilon$.
The algorithm is repeated KL projections to $\mathcal{C}_1,\mathcal{C}_3,\mathcal{C}_3$, together with the computation of an auxiliary sequence $\{q_{k}\}_{k\geq 1}$.

\algrenewcommand\algorithmicrequire{\textbf{Input:}}
\algrenewcommand\algorithmicensure{\textbf{Output:}}

\begin{algorithm}[H]
\caption{Our method}\label{alg:Dykstra}
\begin{algorithmic}
\Require $\supp{X},\supp{\td{X}},N$.
\Require $P^{X},P^{\td{X}},V$, $\Theta,\epsilon,C,K,\varepsilon$.
\State Define three convex sets
$$\mathcal{C}_1=\{\gamma\in\bb{R}^{N\times N}_+\mid\gamma\mathbb{1}=P^{X}\},\;\mathcal{C}_2=\{\gamma\in\bb{R}^{N\times N}_+\mid\gamma^{\tr}\mathbb{1}=P^{\td{X}}\},$$
$$\mathcal{C}_3=\{\gamma\in\bb{R}^{N\times N}_+\mid -\Theta\leq\gamma^{\tr}V\leq\Theta\}.$$
\State Initialise $\gamma^{(0)}=\exp{(-C/\epsilon)}$ \Comment{Dykstra's Algorithm}
\For{$k=1,\dots,3$}
\State Compute $\gamma^{(k)}=\prox^{KL}_{\mathcal{C}_k}(\gamma^{(k-1)})$.
\EndFor
\For{$k=4,\dots,K$}
\State Set $\mathcal{C}_k=\mathcal{C}_{1+(k\mod 3)}$
\State Compute
\begin{equation}
q_{k-3}:=
\begin{cases}
\frac{\gamma^{(k-4)}}{\gamma^{(k-3)}}& \textrm{ if }k=4,\dots,7\\
q_{k-7}\odot\frac{\gamma^{(k-4)}}{\gamma^{(k-3)}}&\textrm{ otherwise}.
\end{cases}\label{equ:q-seuquence}
\end{equation}
\State Compute $\gamma^{(k)}=\prox^{KL}_{\mathcal{C}_{k}}(\gamma^{(k-1)}\odot q_{k-3})$. \Comment{See Lemma~\ref{lem:c_1&c_2}-\ref{lem:prox_c3}}
\EndFor
\Ensure The solution of Equations~(\ref{equ:our-formulation-2}-\ref{equ:convex-sets}): $\gamma^{(K)}$.
\end{algorithmic}
\end{algorithm}

\color{black}
Our computation of KL projections relies on the subgradient optimality condition.
For a set $\mathcal{C}$ in $\bb{R}^{N\times N}$, we define its indicator function $\iota_{\mathcal{C}}$ as:
\begin{equation}
\iota_{\mathcal{C}}(x):=
\begin{cases}
0&\textrm{ if } x\in\mathcal{C},\\
+\infty&\textrm{ otherwise.}
\end{cases}
\end{equation}
Let $\partial\iota_{\mathcal{C}}(\gamma^*)$ denote the set of subgradients $\nu$ of $\iota_{\mathcal{C}}$ at $\gamma^*\in\dom\iota_{\mathcal{C}}$ satisfying Equation~\eqref{equ:subgradient-define}, with ``$\textrm{dom}$'' being the essential domain:
\begin{equation}
\iota_{\mathcal{C}}(\gamma)\geq \iota_{\mathcal{C}}(\gamma^*)+\langle \nu,(\gamma-\gamma^*)\rangle, \forall \gamma\in\bb{R}^{N\times N}.\label{equ:subgradient-define}
\end{equation}
We now introduce this optimality condition.
\begin{lemma}[Subgradient optimality condition of KL projections]\label{lem:subgradient-optimality}
If $\mathcal{C}$ is a closed convex set, the KL projection $\prox^{KL}_{\mathcal{C}}(\bar{\gamma})$ has a unique solution $\gamma^*$, such that
\begin{equation}
\mathbb{0}\in\log\left(\frac{\gamma^*}{\bar{\gamma}}\right)+\partial\iota_{\mathcal{C}}(\gamma^*).
\end{equation}
\begin{proof}
See Appendix~\ref{app:lem:subgradient-optimality}.
\end{proof}
\end{lemma}
\color{black}

Now, we explain how to compute these KL projections in our algorithm:
\begin{lemma}[KL projections for $\mathcal{C}_1,\mathcal{C}_2$]
\label{lem:c_1&c_2}
The KL projection for $\mathcal{C}_1,\mathcal{C}_2$ in Algorithm~\ref{alg:Dykstra},~\ref{alg:baseline} could be computed in closed form:
\begin{equation}
\prox^{KL}_{\mathcal{C}}(\bar{\gamma})=
\begin{cases}
\diag{\frac{P^{X}}{\bar{\gamma}\mathbb{1}}}\bar{\gamma} & \textrm{if } \mathcal{C}=\{\gamma\in\bb{R}^{N\times N}_+\mid\gamma\mathbb{1}=P^{X}\}\\
\bar{\gamma}\diag{\frac{P^{\td{X}}}{\bar{\gamma}^{\tr}\mathbb{1}}} & \textrm{if } \mathcal{C}=\{\gamma\in\bb{R}^{N\times N}_+\mid\gamma^{\tr}\mathbb{1}=P^{\td{X}}\}
\end{cases},
\end{equation}
where the division operator is element-wise.
\begin{proof}
See Appendix~\ref{app:lem:c_1&c_2}.
\end{proof}
\end{lemma}


\color{black}
We define the set of non-zero entries of $V$ as $\overline{\supp{X}}:=\{i\in\supp{X}\mid V_i\neq 0\}$, because we only need to consider $\bar{\gamma}_{i,j}V_i$ when $i \in \overline{\supp{X}}$ (otherwise, it is zero) for the computation of $\prox^{KL}_{\mathcal{C}_3}(\bar{\gamma})$.
\begin{lemma}[KL projections for $\mathcal{C}_3$]
\label{lem:prox_c3}
Under Assumption~\ref{ass:finite}, 
let $\gamma^*:=\prox^{KL}_{\mathcal{C}_3}(\bar{\gamma})$. Then, for all $i\in\supp{X},j\in\supp{\td{X}}$, the entries of $\gamma^*$ are
\begin{equation}
\gamma^*_{i,j}=
\begin{cases}
\bar{\gamma}_{i,j}\exp{(-V_i\nu_j )}\quad &\textrm{if }i\in\overline{\supp{X}}  \textrm{ and } [\bar{\gamma}^{\tr}V]_j\notin[-\Theta_j,\Theta_j] \\
\bar{\gamma}_{i,j} \quad &\textrm{if }i\notin\overline{\supp{X}} \textrm{ or  } [\bar{\gamma}^{\tr}V]_j\in[-\Theta_j,\Theta_j]
\end{cases},\label{equ:proj3_entry}
\end{equation}
where $[\bar{\gamma}^{\tr} V]_j$ is the $j^{th}$ entry of the vector $\bar{\gamma}^{\tr}V$, and the entries of $\nu\in\mathbb{R}^N$ satisfy:
\begin{equation}
\sum_{i\in\overline{\supp{X}}}\bar{\gamma}_{i,j}V_i\exp{(-V_i\nu_j )}=
\begin{cases}
\Theta_j &\textrm{if } [\bar{\gamma}^{\tr}V]_j > \Theta_j\\
-\Theta_j &\textrm{if } [\bar{\gamma}^{\tr}V]_j < -\Theta_j
\end{cases}
,\forall j\in\supp{\td{X}}.
\label{equ:nu_condition}
\end{equation}
\begin{proof}
See Appendix~\ref{app:lem:prox_c3}.
\end{proof}
\end{lemma}
\color{black}

As stated in Lemma~\ref{lem:prox_c3}, the vector $\nu$ is essential for computing $\prox^{KL}_{\mathcal{C}_3}(\bar{\gamma})$.
For an arbitrary $j\in\supp{\td{X}}$, the entry $\nu_j$ in Equation~\eqref{equ:nu_condition} is indeed the root of function $F(x):=\sum_{i\in\overline{\supp{X}}}\bar{\gamma}_{i,j}V_i\exp{(-V_i x)} + c$, where $c=\pm\Theta_j$ is a constant. Its first derivative 
$\nabla F(x)=-\sum_{i\in\overline{\supp{X}}}\bar{\gamma}_{i,j}(V_i)^2\exp{(-V_i x)}\leq 0$,
such that $F(x)$ is non-increasing.
\textcolor{black}{Under Assumption~\ref{ass:finite}, Appendix~\ref{app:rem:nu_j} shows that the root (i.e., $\nu_j$) exists and has finite value.}
Then, $\nu_j$ can be found by root-finding algorithms e.g., Newton-Raphson algorithm, bisection method. 

\color{black}
Although Appendix~\ref{app:rem:nu_j} demonstrates that $\nu_j$ exists and is finite, computational issues can arise with the term $\exp(-V_i\nu_j)$ in the upper case of Equation~\eqref{equ:proj3_entry}, as it may become large enough to be treated as infinity, or small enough to be treated as zero by the computer.
Consequently, the solution for $\prox^{KL}_{\mathcal{C}_3}$ might contain infinite entries and zeros during computation. 
Further, as very small entries in the coupling are considered zero, it may lead to division by zero when calculating the sequence $q_k$ in Equation~\eqref{equ:q-seuquence}, through element-wise division of $\frac{\gamma^{(k-4)}}{\gamma^{(k-3)}}$.
In the following experiments, we terminate iterations if zero-division issues occur and skip the fold if infinite entries exist.

To improve numerical stability, it is preferable to use a relatively larger $\Theta$ (e.g., $(1e-2)\times\bb{1}$ or $(1e-3)\times\bb{1}$), because fewer entries will fall into the upper case of of Equation~\eqref{equ:proj3_entry} when computing $\prox^{KL}_{\mathcal{C}_3}$, and as shown in Figure~\ref{fig:exa_couplings}, the coupling becomes less sparse as $\Theta$ increases, reducing the number of very small entries.

\begin{lemma}[Convergence of Algorithm~\ref{alg:Dykstra}]\label{lem:convergence_algorithm1}
Under Assumption~\ref{ass:finite}, our method in Algorithm~\ref{alg:Dykstra} converges to the optimal solution of Equations~(\ref{equ:our-formulation-2}-\ref{equ:convex-sets}). 
\begin{proof}
Algorithm~\ref{alg:Dykstra} is a special case of Dykstra's algorithm, whose convergence result is explained in Appendix~\ref{app:Dykstra}.
Then, Lemma~\ref{lem:existence} ensures that our formulation in Equations~(\ref{equ:our-formulation-2}-\ref{equ:convex-sets}) satisfies the assumption of Dykstra's algorithm.
\end{proof}
\end{lemma}

Given the vector $\Theta\in\bb{R}^N_+$ and Assumption~\ref{ass:finite} satisfied, Lemma~\ref{lem:convergence_algorithm1} guarantees that Algorithm~\ref{alg:Dykstra} converges to the unique coupling of Equations~(\ref{equ:our-formulation-2}-\ref{equ:convex-sets}).
Then, projecting the source data via the group-blind map induced from the unique coupling, we can achieve $\Theta$-repair in the projected data.
\color{black}

\subsection{Higher dimensions}
\label{sec:higher-dimension}
Now, we consider situations with more than one unprotected attribute:

\paragraph{There are extra attribute(s) $U\in\bb{R}^{d},d\geq 1$ that need not to be adjusted.} In this case, a sample becomes $(i,u,s)$, where $i\in\supp{X}$, and $u,s$ are the values of attributes $U,S$.
As mentioned in Definition~\ref{def:projection}, the projection map $\mathcal{T}$ would not change any attributes other than $X$. 
Hence, this sample is projected into a sequence of weighted samples $\{(j,w_{j},u,s)\}_{j\in\supp{\td{X}}}$. 
    
\paragraph{There are extra attribute(s) that need to be adjusted.} Without loss of generality, we assume attributes $X_1,X_2\in\mathbb{R}$ need to be adjusted. Let $X=(X_1,X_2)$ and $\supp{X}=\supp{(X_1,X_2)}\subseteq\bb{R}^2$, such that $\supp{X}$ includes all tuples $(x_1,x_2)$ present in the source data, where $x_1\in\supp{X_1},x_2\in\supp{X_2}$.
Then $N$ is the cardinality of the set $\supp{X}$. In the same manner, we can define $\supp{\td{X}}$ and the probability distributions:
\begin{equation*}
\begin{split}
P^X_i:=\p{(X_1,X_2)=i}&,\quad
P^{X_s}_i:=\pover{(X_1,X_2)=i}{S=s}, \forall i\in\supp{X},\\P^{\td{X}}_j:=\p{(\td{X}_1,\td{X}_2)=j}&,\quad P^{\td{X}_s}_j:=\pover{(\td{X}_1,\td{X}_2)=j}{S=s}, \forall j\in\supp{\td{X}}.
\end{split}
\end{equation*}
It is worth noting that the cost matrix $C$ needs to consider the range of each attribute that need to be adjusted. Suppose $\supp{X_1}=\{1,2\}$ and $\supp{X_2}=\{1,\dots,4\}$. While not necessarily appropriate in practice, if the entries of cost matrix $C_{i,j}:=\|i-j\|_1$, the cost of moving $X_1$ from $1$ to $2$ would be treated the same as moving $X_2$ from $1$ to $2$. Instead, we can set $C_{i,j}:=\|g\odot (i-j)\|_1$, where $g\in\bb{R}^{2}_+$ denotes the weights for the cost of moving one unit of $X_1,X_2$, and $\odot$ is element-wise product. If $g=[1,1/4]$, the cost of moving $X_1$ from $1$ to $2$ is the same as moving $X_2$ from $1$ to $4$.

\subsection{Choices of target distributions} 
There is no an explicit requirement regarding $P^{\td{X}}\in\Sigma_N$ in Lemma~\ref{lem:convergence_algorithm1}. It leaves the choice of target distribution open.
\begin{itemize}
\item  Concerning with data distortion, one option would be the barycentre distribution, because it brings the least alteration to source data, from its definition in Equation~\eqref{equ:barycentre-define}, or in the original works of \cite{gordaliza2019obtaining,oneto2020fairness,jiang2020wasserstein}. 
In our setting, the requirement regarding the protected attribute $S$ is the vector $V$ only. If those $S$-wise distributions $P^{X_{s_0}},P^{X_{s_1}}$ are not explicitly given, we cannot compute the barycentre between them. But we include it as barycentre baseline in Section~4.2.
\item The other option concerns with performance or utility of some pre-established machine learning models, \textcolor{black}{in the context of transfer learning, where there is a pre-established model trained on large data but on a probably different domain.}
Suppose the classification or prediction model is trained on a training set.
We set the target distribution the same as the the distribution of the training set, or other common representation space, to preserve the classification or prediction performance of the pre-trained model.
\end{itemize}

\section{Numerical Illustrations}

Repair schemes consist of mapping source data into the pre-designed target data, and after the mapping, we obtain the projected data, which are expected to have equalised distributions of attributes in privileged and unprivileged groups.
During the mapping, a sample in the source data is divided into a sequence of weighted samples, leading to data distortion---specifically, a deviation from its original representation.
The data distortion may dampen accuracy or utility of some pre-trained prediction or classification models.
For example, recall the motivating example introduced in Section~1.1. If all applicants are projected to the maximum exam score, the cut-off point loses its ability to distinguish between privileged and unprivileged groups but also significantly forfeits its capacity to classify qualified applicants.

This section first demonstrates the effects of our total repair and partial repair schemes using synthetic data, and then illustrates the classic trade-off between bias repair and data distortion, using the adult census dataset of \cite{misc_adult_2}.
Specifically,
Section~4.1 explains how we use the protected attribute in experiments and introduce the performance indices. Section 4.2 outlines the baselines, while Section 4.3 introduces the real-world datasets that will be discussed in Section 5. Lastly, Section 4.4 describes our experiments conducted on synthetic data, including comparisons with several baselines.
Our implementation is available online \footnote{\url{https://github.com/Quan-Zhou/OT_Debiasing}}.



\subsection{Input and Indices}

We have made the assumption that the protected attribute $S$ of each sample is not observed, but the vector $V$, as defined in Theorem~\ref{pro:binary_condition}, is given as the input for our algorithm.
The computation of $V$ requires the population-level information regarding the distributions of attributes for both groups ($s_0$ and $s_1$), given the crucial assumption that the source data represents unbiased samples from the broader population, such that groups in source data follow the same distributions.
In our experiments, since there is no such population-level information given, we directly compute $V$ from the source data. 
Apart from computing $V$, the protected attribute of each sample is not utilised for computing the coupling or for projecting source data in our method.


Next, to introduce the performance indices, we start from defining computation of empirical distributions:
\begin{definition}[Empirical distributions]\label{def:empirical}
Let $\supp{Z}\in\bb{R}$.
Given data $(i,w_{i})_{i\in\supp{Z}}$, the empirical distribution $P^Z$ is defined as
\begin{equation}
P^Z_i:=\frac{w_i}{\sum_{i\in\supp{Z}} w_{i}}.
\label{equ:PY-define}
\end{equation}
For instance, Equation~\eqref{equ:PY-define} computes $P^{X}$ if $Z=X$, and $P^{\td{X}}$ if $Z=\td{X}$.
If the  protected attribute $S$ is known, such that the data become $(i,s,w_{i,s})_{i\in\supp{Z},s\in\supp{S}}$, we can compute the $S$-wise distributions $P^{Z_{s}}$ as follows:
\begin{equation}
P^{Z_{s}}_i:=\frac{w_{i,s}}{\sum_{i\in\supp{Z}} w_{i,s}},\;\forall\; s\in\supp{S}.
\label{equ:PY|s-define}
\end{equation}
\end{definition}

\begin{definition}[Performance indices]\label{def:performance}
Let $Y$ and $f$ denote the label that we need to estimate and the estimated label.
To measure the performance of prediction / classification models on the source or projected data, we introduce f1 scores as accuracy measures, disparate impact as the fairness measure, and $S$-wise TV distance:
\begin{align*}
\textrm{f1 micro}:=&\frac{\sum_{s\in\supp{S}} (2\times TP_s)}{\sum_{s\in\supp{S}} (2\times TP_s + FP_s + FN_s)},\\
\textrm{f1 macro}:=&\sum_{s\in\supp{S}}\frac{1}{2}\times\frac{2\times TP_s}{2\times TP_s + FP_s + FN_s},\\
\textrm{f1 weighted}:=&\sum_{s\in\supp{S}}{P^{S}_s}\times\frac{2\times TP_s}{2\times TP_s + FP_s + FN_s},\\
\textrm{DI}:=&\frac{\pover{f=1}{S=s_0}}{\pover{f=1}{S=s_1}},\\
S\textrm{-wise TV distance}:=&\tv{P^{\td{X}_{s_0}}}{P^{\td{X}_{s_1}}},
\end{align*}
where $s_0$ denotes the unprivileged group and $Y=0$ is the negative label.
$TP_s,FP_s,FN_s$ are the numbers of true positive, false positive and false negative of the group $s$.
Empirical distributions $P^{S},P^{\td{X}_{s_0}},P^{\td{X}_{s_1}}$ are computed from the projected data, following Definition~\ref{def:empirical}.
\end{definition}
Ideally, after the repair, f1 scores increase, disparate impact gets closer to $1$, and $S$-wise TV distance decreases.


\subsection{Baselines}
\label{sec:baselines}

Now, we have selected a range of established baselines that reflect the current practices in the field.

\textcolor{black}{\paragraph{Origin} This baseline keeps the data unchanged.}

\paragraph{Unconstrained OT}
This baseline (abbreviated as ``unconstrained'') is to solve the standard regularised OT:
\begin{equation}
\min_{\gamma\in\Pi(P^{X},P^{\td{X}})} \langle C,\gamma \rangle -\epsilon E(\gamma)=\kl{\gamma}{\xi}, \textrm{ where }\Pi(P^{X},P^{\td{X}})=\bigcap^{2}_{\ell=1}\mathcal{C}_{\ell},\;\xi=\exp{(-C/\epsilon)}.\label{equ:formulation-baseline}
\end{equation}
The only difference between this baseline and our method is that we add one additional constraint $-\Theta\leq\gamma^{\tr}V\leq\Theta$.
The optimal solution of Equation~\eqref{equ:formulation-baseline} can be founded by Dykstra's Algorithm, or computationally cheaply using the iterative Bregman projections, as suggested in Algorithm~\ref{alg:baseline}. 
Both algorithms converge at the same results, because $\mathcal{C}_1,\mathcal{C}_2$ are affine subspaces (See Fact~1.4 in \cite{bauschke2020dykstra}, or Theorem~4.3 in \cite{bauschke2000dykstras}).

\begin{center}
\begin{algorithm}[H]
\caption{Unconstrained OT method}\label{alg:baseline}
\begin{algorithmic}
\Require $\supp{X},\supp{\td{X}}, N$, $P^{X},P^{\td{X}}$, $\epsilon,C,K$.
\State Set $$\mathcal{C}_1=\{\gamma\in\bb{R}^{N\times N}_+\mid\gamma\mathbb{1}=P^{X}\},\;\mathcal{C}_2=\{\gamma\in\bb{R}^{N\times N}_+\mid\gamma^{\tr}\mathbb{1}=P^{\td{X}}\}.$$
\State Initialise $\gamma^{(0)}=\exp{(-C/\epsilon)}$. \Comment{Iterative Bregman Projections}
\For{$k=1,\dots,K$}
\State Set $\mathcal{C}_k=\mathcal{C}_{1+(k\mod 2)}$
\State Compute $\gamma^{(k)}=\prox^{KL}_{\mathcal{C}_k}(\gamma^{(k-1)})$, using Lemma~\ref{lem:c_1&c_2}.
\EndFor
\Ensure The solution of Equation~\eqref{equ:formulation-baseline}: $\gamma^{(K)}$.
\end{algorithmic}
\end{algorithm}
\end{center}

\paragraph{The barycentre projection in \cite{gordaliza2019obtaining}}
This baseline (abbreviated as ``barycentre'') is the total repair in Section~5.1.1 (B) of  \cite{gordaliza2019obtaining}, where each sample split its mass to be transported, the same as our setting.
The idea is that the two conditional distributions of the random variable $X$ by the protected attribute $S$ are going to be transformed into the Wasserstein barycentre $P^{B}$ between $P^{X_{s_0}}$ and $P^{X_{s_1}}$, with weight $\pi_0$ and $\pi_1$. The 1-Wasserstein barycentre is defined as
\begin{equation}
P^{B}=\arg\min_{P\in\Sigma_{N}}\{\pi_0 \mathcal{W}_1(P^{X_{s_0}},P)+\pi_1 \mathcal{W}_1(P^{X_{s_1}},P)\},
\label{equ:barycentre-define}
\end{equation}
where $\mathcal{W}_1(P^{X_{s_0}},P^{X_{s_1}}):=\sum_{i\in\supp{X}} |P^{X_{s_0}}_i - P^{X_{s_1}}_i|$ denotes the $1$-Wasserstein distance between two distributions $P^{X_{s_0}},P^{X_{s_1}}$ defined on $\supp{X}$. 
The non-negative constants $\pi_0+\pi_1=1$ are barycentric coordinates.
The barycentre within two marginal distributions can be found via the optimal coupling between $P^{X_{s_0}},P^{X_{s_1}}$ (Remark~4.1 in \cite{gordaliza2019obtaining}, or Proposition~5.9 in \cite{villani2021topics}). 
The implementation is to find the coupling $\gamma^B\in\Pi(P^{X_{s_0}},P^{X_{s_1}})$, \textcolor{black}{as in Algorithm~\ref{alg:baseline} with $X,\td{X}$ replaced by $X_{s_0},X_{s_1}$}, and then transport the group of $s_0$ by the map corresponding to the coupling $\gamma^{0\to B}$ and the group of $s_1$ by the map corresponding to the coupling $\gamma^{1\to B}$, whose entries are defined by 
\begin{equation}
\gamma^{0\to B}_{i,k}:=\gamma^{B}_{i,\;\pi_0\cdot i + \pi_1\cdot k},\quad \gamma^{1\to B}_{i,k}:=\gamma^{B}_{\pi_1\cdot i + \pi_0\cdot k,\; i}.
\end{equation}


\color{black}
Below, we present several baselines that involve adjusting attributes, labels, predicted labels, or combinations thereof to support demographic parity (cf. Section~\ref{sec:related-work}) as the fairness notion. These methods are implemented using the AI Fairness 360 toolkit\footnote{available at \url{https://github.com/Trusted-AI/AIF360}} \citep{aif360-oct-2018}, an extensible open-source library containing techniques developed by the research community to help detect and mitigate bias in machine learning models throughout the AI application lifecycle.

\paragraph{Reject option classification in \cite{kamiran2012decision}} This baseline (abbreviated as ``ROC'') is a post-processing technique that adjusts predicted labels in favour of unprivileged groups and against privileged groups within a confidence band around the decision boundary, where predictions are most uncertain \citep{kamiran2012decision}. The optimal classification threshold and margin are estimated on a validation set to optimise a fairness metric  (here set as ``statistical parity difference'', equivalent to mitigate disparate impact). These thresholds are then used to revise predicted labels on test set. This method requires the protected attribute in both the validation and test sets.

\paragraph{Learning fair representations in \cite{zemel2013learning}} 
This baseline (abbreviated as ``LFR'') is a disparate impact mitigation method that transforms the unprotected attribute into a latent representation, which encodes the data well while obfuscating information about protected attributes \citep{zemel2013learning}.
The transformation map first projects the unprotected attribute into the latent representation and then directly predicts the label based on this representation. 
This map is group-blind, but training it requires the protected attribute. 
The parameters we used in training are $Ax = 0.01$ and $Ay = Az = 1$, whereas in the original paper, $Ax$ was $0.01$ and $Ay,Az$ were chosen from the set ${0.1, 0.5, 1, 5, 10}$.
We train the map using the training set and apply it to the test set, meaning the protected attribute is required in the training set but not in the test set.

\paragraph{Disparate impact remover in \cite{feldman2015certifying}}
This baseline (abbreviated as ``DIremover'') is a preprocessing technique that edits unprotected attribute to 
align the distributions conditioned on the protected attribute while preserving rank-order within groups \citep{feldman2015certifying} in the given dataset. We transform both the training and test sets, with the repair level set to the maximum value of $1.0$. The protected attribute is required in both.

\paragraph{Reweighing in \cite{kamiran2012data}}
This baseline (abbreviated as ``RW'') is a preprocessing technique that weights the samples in each (group, label) combination differently to ensure statistical
parity before classification \citep{kamiran2012data}. We transform only the training set, where the protected attribute is required.
\color{black}

\subsection{Datasets}
\paragraph{Adult dataset in \citep{misc_adult_2}.}
The adult census dataset\footnote{Downloaded from \url{https://archive.ics.uci.edu/dataset/2/adult}}, comprises of 48842 samples of 14 attributes (e.g., sex, race, age, education level, marital-status, occupation) and a high-income indicator.
The high-income indicator denotes whether the annual income of a sample is lower than \$50K (label $Y=0$) or higher (label $Y=1$).
Of the 14 attributes, five are numerical: ``education-num'' (ranging from 1 to 16), ``hours-per-week'', ``age'', ``capital-gain'', ---and all are selected as unprotected attributes. \textcolor{black}{These numeric attributes, except for ``education-num'', are mapped to integers ranging from 0 to 4 for two reasons: 1) their distributions are highly skewed. For example, ``capital-gain'' ranges from 0 to 99,999, yet over $90\%$ of the samples have values below 100; 2) the computational complexity of our method, ``unconstrained'', ``barycentre'' depends on the number of discretisation points $N$. In higher dimensions, $N$ is the product of the discretisation points of each attribute. Although ``age'' and ``hours-per-week'' are discrete, each has roughly 100 discretisation points.
Then, mapping attributes to the 0-4 range produces the \textit{cleaned Adult dataset}, referred to simply as the \textit{Adult dataset} hereafter.}
There are two commonly-recognised protected attributes, i.e., ``race'' and ``sex''.

If the protected attribute $S$ is ``race'', we select all samples ($M=46447$) whose race is ``white'' ($S=1$) or ``black'' ($S=0$).
If $S$ is ``sex'', we select all samples ($M=48842$) whose gender is female ($S=0$) or male ($S=1$).
For each numerical attribute, we compute the TV distance between race-wise (resp. gender-wise) marginal distributions in Table~\ref{tab:adult-TV}.
\textcolor{black}{
They serve as a reference to identify which attributes need adjustment. We focus specifically on proxies of the protected attribute, as these allow group-blind models to distinguish between privileged and unprivileged groups, leading to disparate impact.}
We let $X$ include all attributes with $S$-wise TV distance higher than $0.1$. Specifically, $X$ contains ``education-num'', ``hours-per-week'' if $S$ is ``race'', and also includes``age'' if $S$ is ``sex''. The other attributes, denoted as $U$, remain unadjusted.
\begin{table}[!htp]
\centering
\begin{tabular}{|l|c|c|c|}
\hline
\textbf{Attributes} & \textbf{Race-wise TV dist.}&\textbf{Gender-wise TV dist.} &\textbf{Importances} \\\hline
age              & 0.0415&\textbf{0.1010} & 0.2006\\
education-num    & \textbf{0.1187}& 0.0710 & 0.3177\\
capital-gain      & 0.0268&0.0369 & 0.3416 \\
capital-loss      & 0.0142&0.0201 & 0.0476\\
hours-per-week     & \textbf{0.1222}&\textbf{0.1819} & 0.0924\\\hline
\end{tabular}
\caption{
The first two columns present the Total Variation (TV) distances between race-wise and gender-wise marginal distributions for five numerical attributes in the \textcolor{black}{cleaned} Adult dataset \citep{misc_adult_2}. Attributes with a TV distance greater than 0.1, highlighted in the table, are selected as set $X$ and will undergo adjustment. The remaining attributes are labelled as $U$. \textcolor{black}{The ``importances'' column displays the mean of attribute Gini importances in trained random-forest models across a 10-fold cross-validation.}}
\label{tab:adult-TV}
\end{table}

\color{black}
\paragraph{COMPAS dataset in \cite{angwin2022machine}.}
The COMPAS (Correctional Offender Management Profiling for Alternative Sanctions) dataset\footnote{Downloaded from \url{https://raw.githubusercontent.com/propublica/compas-analysis/master/compas-scores-two-years.csv}} is a widely used dataset in fairness research, specifically within the context of criminal justice. It includes data on individuals assessed by the COMPAS risk assessment tool, which predicts the likelihood of a defendant reoffending. 
The dataset includes attributes of age, sex, race, prior criminal history, charge degree, and a recidivism label indicating whether an individual reoffended $(Y=1)$ or not $(Y=0)$ within two years of the initial assessment.

As it was originally collected as an investigation into potential racial biases in COMPAS scores, we consider ``race'' as the protected attribute.
We select all samples ($M=6172$) whose race is ``Caucasian'' (S = 1) or ``Not Caucasian'' (S = 0). For unprotected attributes, we use all other attributes, excluding ``sex'' (as it is sometimes considered protected attributes) and ``c\_charge\_desc''  (as it is a descriptive, non-numeric attribute).
Similarly, we compute the TV distance between race-wise marginal distributions in Table~\ref{tab:compas-TV}.
We let $X$ include all attributes with $S$-wise TV distance higher than $0.1$.
\begin{table}[!htp]
\centering
\begin{tabular}{|l|c|l|c|}
\hline
\textbf{Attributes} & \textbf{TV distance}&\textbf{Attributes} & \textbf{TV distance}\\\hline
 juv\_fel\_count  & 0.0321 &
  juv\_misd\_count  & 0.0432\\
  juv\_other\_count & 0.0218 &
  priors\_count & \textbf{0.1262}\\
  c\_charge\_degree=F & 0.0784 &
  c\_charge\_degree=M & 0.0784 \\
  age\_cat=Less than 25  & 0.0808 &
  age\_cat=25 - 45 & 0.0544  \\
age\_cat=Greater than 45 & \textbf{0.1352}& & \\\hline
\end{tabular}
\caption{The TV distance between race-wise marginal distributions of nine unprotected attributes in the COMPAS dataset \citep{angwin2022machine}. Highlighted attributes, whose $S$-wise TV distance higher than $0.1$ are chosen as $X$ and would be adjusted. The rest are denoted as $U$.}
\label{tab:compas-TV}
\end{table}

\color{black}

\subsection{Revisiting the motivating example}
In the first example, we compare with the baseline of ``unconstrained'' on synthetic data, to demonstrate that an arbitrary $S$-blind projection map would not reach total repair.

Recall the motivating example introduced in Section~1.1, where there is only one unprotected attribute (the exam score) and one binary protected attribute.
Let $\supp{X}=\supp{\td{X}}=\{-30,-29,\dots,10\}$. Suppose $\supp{S}=\{s_0,s_1\}$ and $\p{s_0}=0.7$, $\p{s_1}=0.3$. 
The random variable $X_{s_0}$ (resp. $X_{s_1}$) follow a discretised Gaussian distribution $\mathcal{N}(-10,6^2)$ (resp. $\mathcal{N}(1,3^2)$).

\begin{figure}[htp]
\centering\includegraphics[scale=0.5]{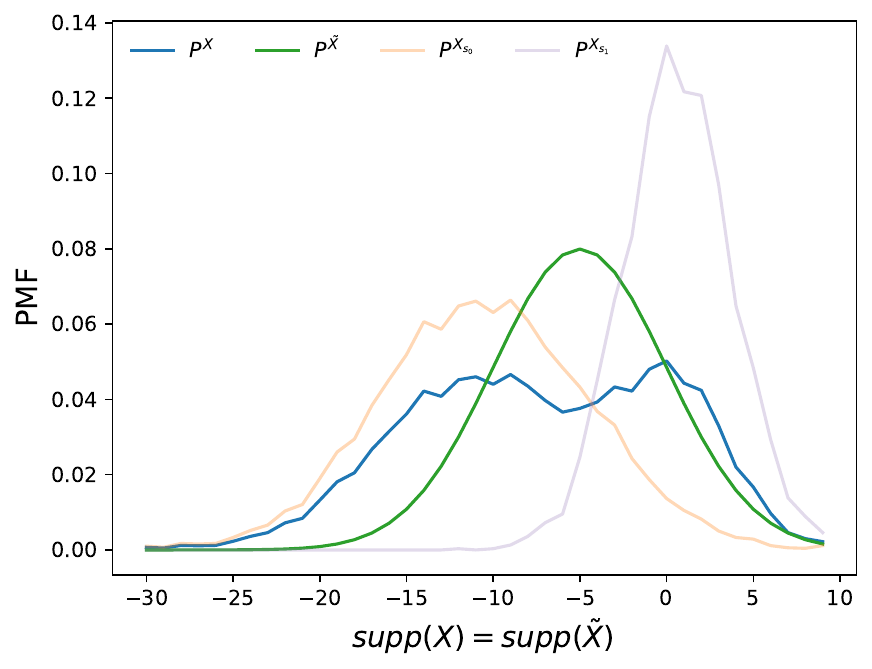}
\caption{Overview of the empirical distributions of $P^{X},P^{X_{s_0}},P^{X_{s_1}}$ from the generated source data and the target distribution $P^{\td{X}}$ that we set arbitrarily.}
\label{fig:exa_overview}
\end{figure}

We generate $M=10^{4}$ samples (source data) by repeating the procedure for $M$ times: first, we generate a uniform random variable within $[0,1]$. If this uniform variable is smaller than $\p{s_0}$, we generate a Gaussian random variable $X_{s_0}\sim\mathcal{N}(-10,6^2)$ and the sample reads $(\lfloor X_{s_0}\rfloor,s_0)$, where $\lfloor \cdot\rfloor$ is the floor function. Otherwise, the sample should be $(\lfloor X_{s_1}\rfloor,s_1)$, where $X_{s_1}\sim\mathcal{N}(1,3^2)$. 
Then, we compute the empirical distributions of $P^{X},P^{X_{s_0}},P^{X_{s_1}},P^{S}$ from the source data.
For the target distribution, we arbitrarily set $P^{\td{X}}$ to the distribution of discretised Gaussian distribution $\mathcal{N}(-5,5^2)$.
In Figure~\ref{fig:exa_overview}, the discrete distributions $P^{X},P^{X_{s_0}},P^{X_{s_1}},P^{\td{X}}$ are displayed by blue, orange, purple, green curves, respectively.

\begin{figure}[htb]
\begin{minipage}{\textwidth}
\begin{minipage}{.24\textwidth}
\centering
\includegraphics[scale=0.5]{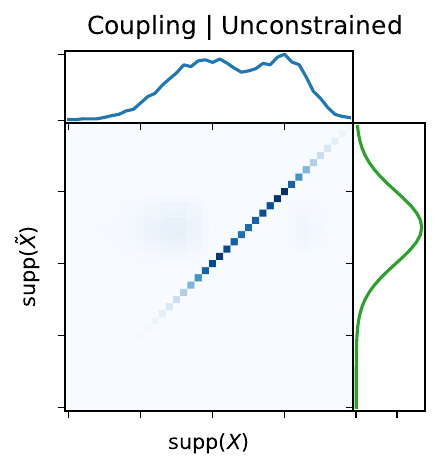}
\label{fig:prob1_6_2}
\end{minipage}%
\begin{minipage}{0.24\textwidth}
\centering
\includegraphics[scale=0.5]{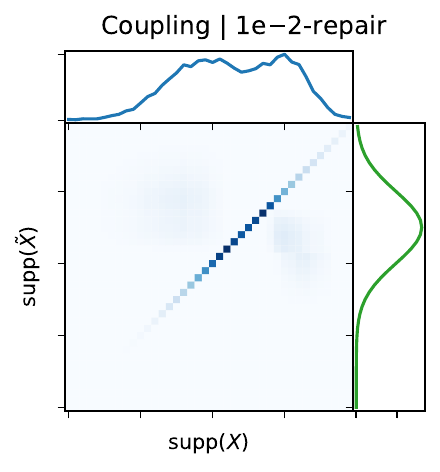}
\label{fig:prob1_6_1}
\end{minipage}
\begin{minipage}{.24\textwidth}
\centering
\includegraphics[scale=0.5]{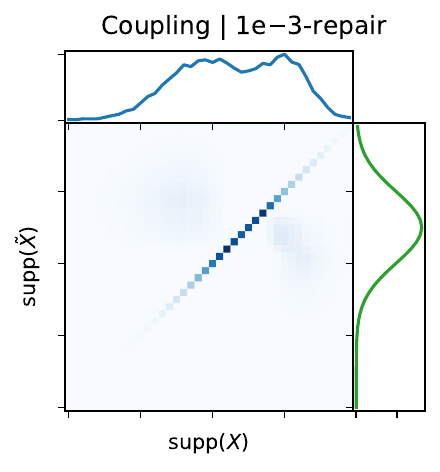}
\label{fig:prob1_6_3}
\end{minipage}%
\begin{minipage}{0.24\textwidth}
\centering
\includegraphics[scale=0.5]{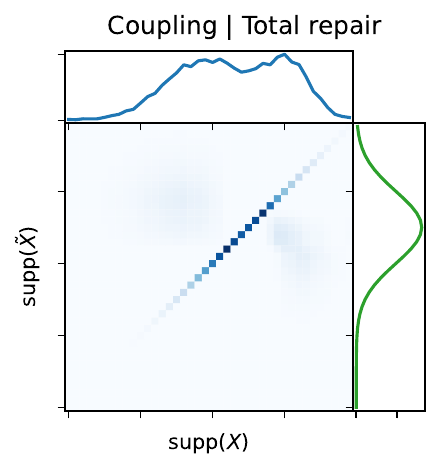}
\label{fig:prob1_6_4}
\end{minipage}
\caption{From left to right: the baseline coupling of Equation~\eqref{equ:formulation-baseline}, 
the optimal coupling $\gamma^*$ of Equations~(\ref{equ:our-formulation-2}-\ref{equ:convex-sets}), when $\Theta=10^{-2}\mathbb{1},10^{-3}\mathbb{1},\mathbb{0}$. The blue and green curves represent marginal distributions $P^{X},P^{\td{X}}$ that are the same across all couplings.}
\label{fig:exa_couplings}
\end{minipage}

\begin{minipage}{\textwidth}
\centering
\includegraphics[width=0.9\textwidth]{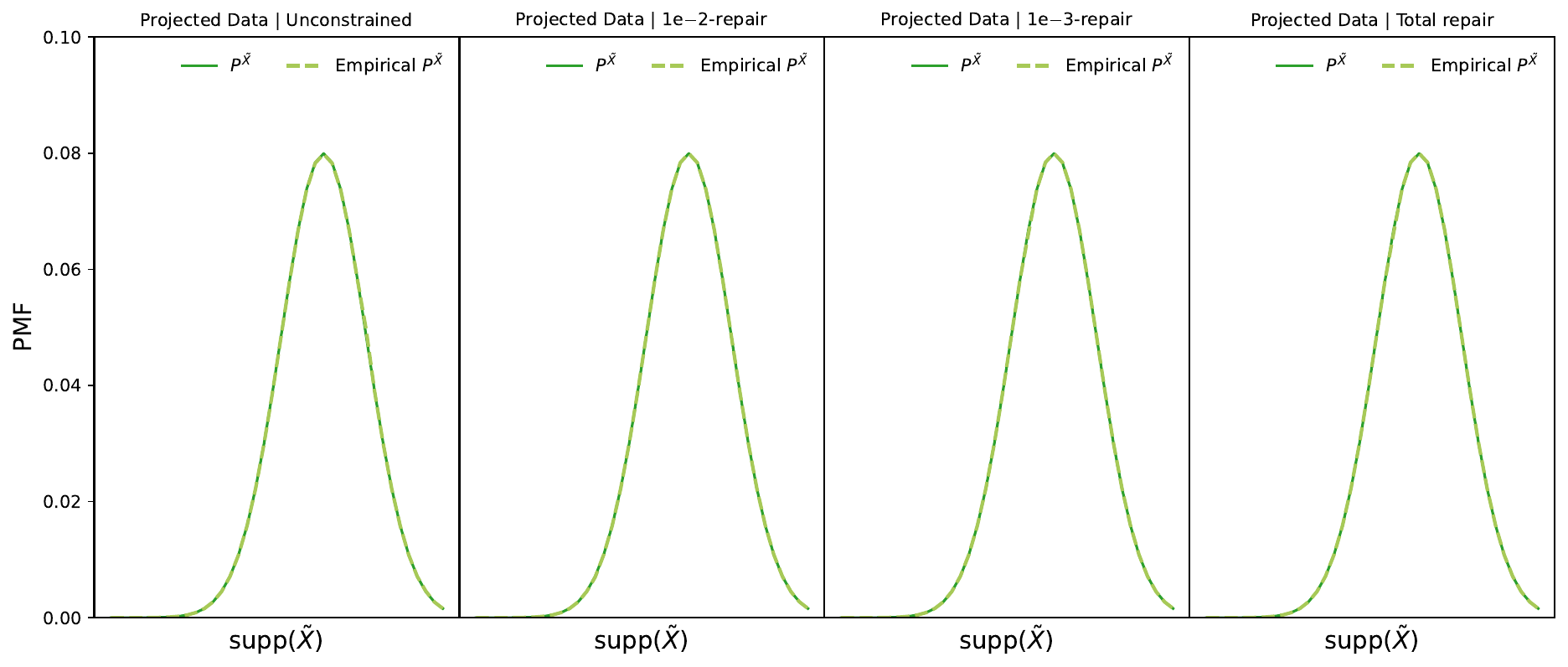}
\caption{\textbf{Group-blind distributions.}
Solid green curves are the $S$-blind target distributions $P^{\td{X}}$ used to compute couplings, which are the green curves in Figure~\ref{fig:exa_couplings}.
Dashed green curves (from left to right) are the $S$-blind empirical distributions of $P^{\td{X}}$ computed from projected data from baseline, from $10^{-2}\mathbb{1}$-repair, from $10^{-3}\mathbb{1}$-repair, and from total repair.
The overlap between dashed curves and solid curves shows the projected data follow the target distribution we design, and verifies that all couplings in Figure~\ref{fig:exa_couplings} are feasible and the projection method in Definition~\ref{def:projection} is correct.}
\label{fig:exa_results_groupbline}
\end{minipage}
\end{figure}

Compute $V=(P^{X_{s_0}}-P^{X_{s_1}})/P^{X}$, set the entry of cost matrix $C_{i,j}=|i-j|$, $\epsilon=0.01$, $\varepsilon=1e^{-4}$ and the number of iteration $K=400$ for baseline and  $K=600$ for our method. Along with $P^X,P^{\td{X}},V,\epsilon,C,K$ and $\Theta=10^{-2}\mathbb{1},10^{-3}\mathbb{1},\mathbb{0}$ as input in Algorithm~\ref{alg:Dykstra}, we find the optimal coupling $\gamma^*$ of Equations~(\ref{equ:our-formulation-2}-\ref{equ:convex-sets}) 
and another coupling of Equation~\eqref{equ:formulation-baseline}, shown in Figure~\ref{fig:exa_couplings}, where the blue and green curves represent $P^{X}$ and $P^{\td{X}}$. 

\begin{figure}[htp]
\centering
\includegraphics[width=0.9\textwidth]{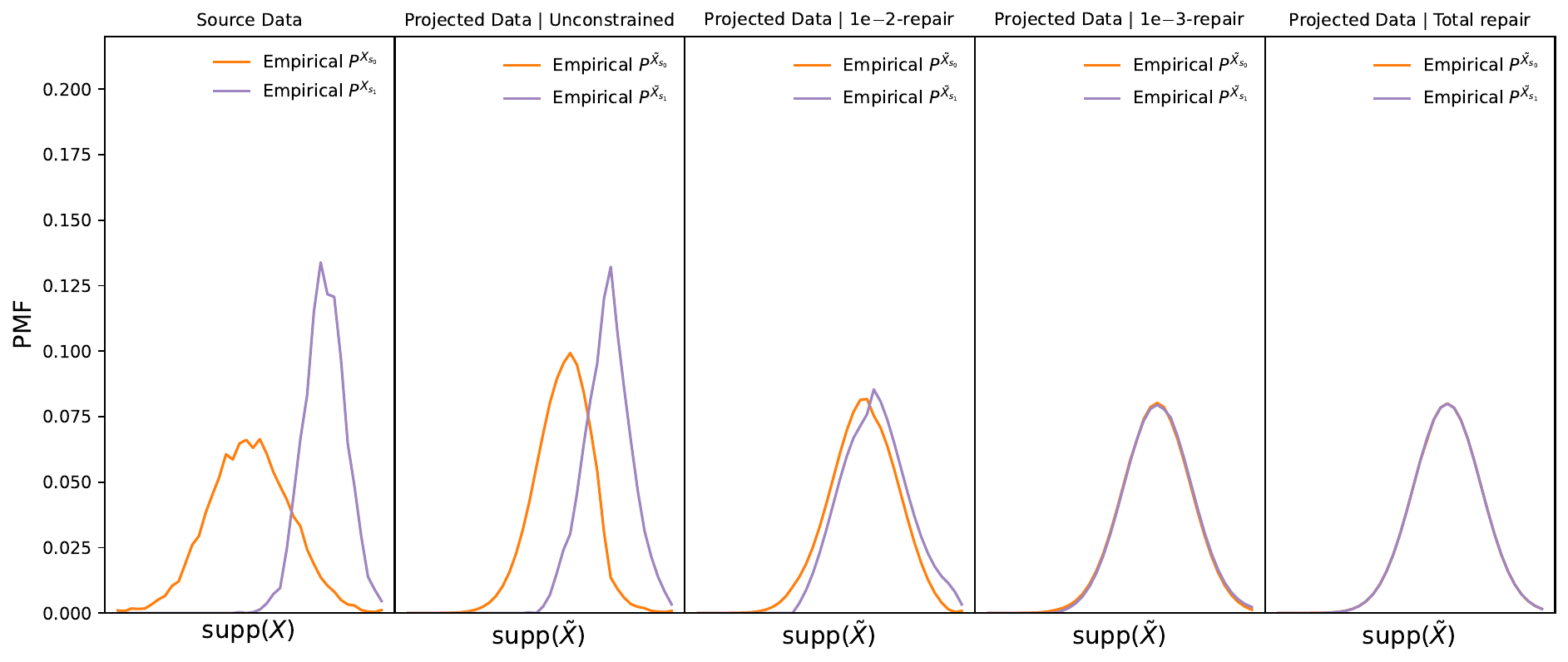}
\caption{\textbf{Group-wise distributions.} From left to right: the $S$-wise empirical distributions of $P^{X_{s_0}},P^{X_{s_1}}$ in source data, $P^{\td{X}_{s_0}},P^{\td{X}_{s_1}}$ in  projected data from baseline, from $10^{-2}\mathbb{1}$-repair, from $10^{-3}\mathbb{1}$-repair, and from total repair. $P^{\td{X}_{s_0}}$ is plotted orange and $P^{\td{X}_{s_1}}$ is plotted purple.}
\label{fig:exa_results}
\end{figure}

Then, for each coupling, we define a projection map following Definition~\ref{def:projection}. 
After applying the projections to source data, we compute the empirical distributions of $P^{\td{X}_{s_0}}$ and $P^{\td{X}_{s_1}}$ from the projected data. 
In Figure~\ref{fig:exa_results}, from left to right, we show the $S$-wise empirical distributions of source data, projected data from baseline, from $10^{-2}\mathbb{1}$-repair, from $10^{-3}\mathbb{1}$-repair, and from total repair, with $P^{\td{X}_{s_0}}$ plotted orange and $P^{\td{X}_{s_1}}$ plotted purple.
We can see that when $\Theta$ gets closer to $\mathbb{0}$, the gap between purple and orange curves shrinks.

In order to demonstrate that all couplings in Figure~\ref{fig:exa_couplings} are feasible, we present the empirical $P^{\td{X}}$ (dashed green curves) computed from projected data and the input of target distribution $P^{\td{X}}$ used to compute all couplings (solid green curves) in Figure~\ref{fig:exa_results_groupbline}. The overlap of both green curves implies that source data are successfully projected to the target data we expect, regardless of the coupling used.



\color{black}
\section{Empirical Analysis of Disparate Impact Mitigation Methods}\label{sec:trade-off}

We conduct random-forest classification on the Adult dataset \citep{misc_adult_2} and COMPAS dataset \citep{angwin2022machine}, to showcase the trade-off between bias repair and data distortion, in comparison with all baselines in Section~\ref{sec:baselines}. Our method, ``unconstrained'', ``barycentre'' are used as post-processing methods.

We performed $10$-fold cross-validation, where, for each fold, we randomly split the data into training, test, and validation sets in a 4:4:2 ratio.
For all baselines except ``LFR'', we used the following procedure.
\begin{itemize}
    \item The training set is transformed if using a preprocessing method (i.e., ``DIremover'', ``RW''); otherwise, it remains unchanged. The (transformed) training set, without the protected attribute, is then used to train a fairness-unaware random-forest model.
    \item For the test set, the attributes in $X$ are transformed if using our method, ``unconstrained'', ``barycentre'' and all the unprotected attributes are transformed if using ``DIremover''; otherwise, they remain unchanged. The trained random-forest model is used to predict the labels.
    \item Predictions on the test set are further transformed if using a post-processing method (i.e., ``ROC'', our method, ``unconstrained'', ``barycentre'').
   \item The validation set is used by ``ROC'' to optimise the classification threshold and by our method, ``unconstrained'' and ``barycentre'' to determine a threshold $f_{th}$, which will be explained in the next paragraph.
\end{itemize}
The baseline ``LFR'' trains a transformation map on the training set, directly applying it to the test set to transform attributes and predict labels without a random-forest model.

In our method, or baselines ``unconstrained''  ``barycentre'', a sample $(x,u,s,y)$ of test set, is split into a sequence of weighted samples $\{(\tx,u,s,y,w_{\tx})\}_{\tx\in\supp{\td{X}}}$. 
Let the random-forest model denoted by $\mathcal{M}$, and the prediction for a single sample is $f_{\tx}=\mathcal{M}(\tx,u)$.
In this case, the repaired prediction would be $0$ if $\sum_{\tx\in\supp{\td{X}}} w_{\tx} f_{\tx}<f_{th}$ and $1$ otherwise.
The threshold $f_{th}$ is set to $0.05$ for both datasets, chosen via grid search on the validation set.
The number of iterations is $K=400$ for ``unconstrained'', ``barycentre'', and is $K=600$ for our method.
The entropic regularisation parameter is $\epsilon=0.01$. 
The entry of cost matrix is $C_{i,j}:=\|g\odot(i-j)\|_1$, where the entry $g_i$ is the reciprocal of the range of the $i^{th}$ attribute in $X$, as discussed in Section~\ref{sec:higher-dimension}.
In our method and ``unconstrained'', the target distribution is simply $P^{\td{X}}=P^X$ to avoid unnecessary data distortion.
The parameter $\pi_0$ of ``barycentre''is the portion of $s_0$ group in source data, the same setting as \cite{gordaliza2019obtaining}.

We measure performance using the disparate impact of the (transformed) training set and (transformed) predictions, f1 scores, and the $S$-wise TV distance of (transformed) attributes $X$ in the test set, as in Definition~\ref{def:performance}.
We divide all methods into two groups. \begin{itemize} 
\item The first group includes methods that transform the training set (i.e., ``LFR'', ``DIremover'', ``RW''), potentially altering the disparate impact of the transformed training set from ``origin''. 
\item The second group includes post-processing methods (i.e., our method, ``unconstrained'', ``barycentre'', ``ROC''), where, except for ``ROC'', the other methods that transform attributes $X$ make the index of the $S$-wise TV distance informative.
\end{itemize}
For verification, we include ``origin'' in both groups, demonstrating identical performance in both, which confirms that the experimental procedures remain the same across groups, with only the methods differing.

\begin{figure}
\centering
\includegraphics[width=\linewidth]{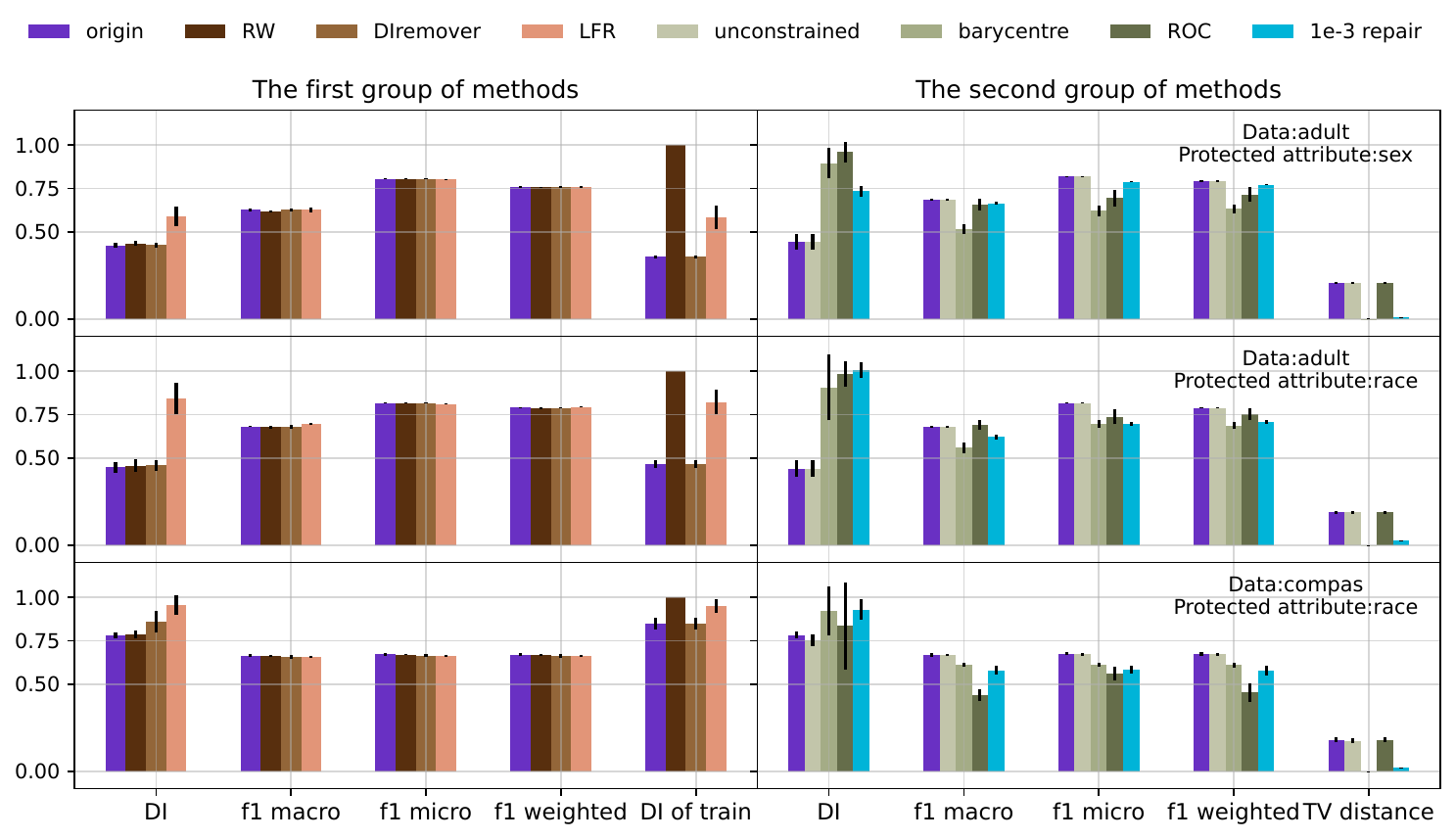}
\caption{\textcolor{black}{Prediction performance for Adult dataset \citep{misc_adult_2} (upper \& middle rows) and COMPAS dataset \citep{angwin2022machine} (lower row).
with the protected attribute being ``sex'' (upper row) and ``race'' (middle \& lower rows). Bars show the mean performance indices, and dark vertical lines indicate mean $\pm$ one standard deviation across $10$-fold cross-validation. 
The left column shows baselines that include preprocessing steps, while the right column displays post-processing methods.
Our partial repair scheme ``1e-3 repair'' is shown in blue. Ideally, disparate impact (``DI'') should be close to $1$, and f1 scores should be as high as possible.}}
\label{fig:E2}
\end{figure}

In Figure~\ref{fig:E2}, we present the disparate impact of the (transformed) training set (labelled as ``DI of train'') and (transformed) predictions (labelled as ``DI''), along with F1 scores (``f1 macro,'' ``f1 micro,'' ``f1 weighted'') and the $S$-wise TV distance of the projected unprotected attribute $X$ (``TV distance'') for all methods across 10-fold cross-validation. The bars and vertical lines represent the mean values and mean $\pm$ one standard deviation, respectively.
The left column displays the first group of methods, and the right column shows the second group. The rows correspond to different datasets and protected attributes: the first row shows results on the Adult dataset with ``sex'' as the protected attribute, the second row with ``race'' as the protected attribute, and the third row on the COMPAS dataset with ``race'' as the protected attribute.

First of all, the identical performance of ``origin'' (in terms of ``DI'' and f1 scores) across both groups of methods confirms that the procedures are consistent, with only the methods differing.

In the left column of Figure~\ref{fig:E2}, ``RW'' significantly improves the ``DI of train'' after re-weighting the training set, but the effect on bias mitigation of prediction is minimal, probably because ``RW'' increases weights for training samples with an unprivileged protected attribute and favourable label, but these weights remain unchanged in the test set. We did not apply ``RW'' to transform the test set, as re-weighting a test sample does not affect its prediction.
The baseline ``DIremover``, shows the same performance of``DI of train'' as ``origin'' as it only transforms unprotected attribute, while this index measures the relationship between the protected attribute and label in the (transformed) training set. Notably, ``DIremover'' has no impact on attribute values for the Adult dataset but does affect those in the COMPAS dataset. This may be due to more distinct group-wise distributions of unprotected attributes (including both``X'' and ``U'') in the COMPAS dataset, where more attributes are retained.
``LFR'' performs well, significantly improving ``DI'' in predictions with minimal accuracy loss, though it shows relatively high variance in bias mitigation performance.

In the right column of Figure~\ref{fig:E2}, ``unconstrained'' performs identically to origin,'' as expected, since there are no constraints promoting fairness.
The baseline `barycentre'' effectively aligns the group-wise distributions of $X$, resulting in a nearly zero ``TV distance''.
However, its bias mitigation performance is unstable, showing a larger accuracy drop when applied to the Adult dataset, likely due to a mismatch between the target and training distributions of the random-forest model. Nonetheless, this method restricts the target distribution to a barycentre.
The ``ROC'' method similarly shows instability, and a significant accuracy decrease when used with the COMPAS dataset, even after we fine-tuned its parameter using the validation set in each fold.
The blue bars ``1e-3 repair'' represent our method with the parameter $\Theta=(1e^{-3})\times\mathbb{1}$.
It demonstrates comparable bias mitigation performance to ``LFR'' but offers greater stability with relatively minor accuracy sacrifices. Importantly, the computation of our projection map does not require the protected attribute and label of each sample, unlike ``LFR''.

Readers might notice that in the first row of Figure~\ref{fig:E2}, the ``DI'' performance of ours are worse than in the second row, despite both using the same Adult dataset.
To investigate this, we computed the Gini importance\footnote{It is measured as the normalised total reduction of the Gini-impurity criterion contributed by each attribute.} for each attribute in the random-forest model trained on the Adult dataset using a 10-fold validation. The mean importance is displayed in the right column of Table~\ref{tab:adult-TV}, where we observe that ``age'', ``education-num'', and ``capital-gain'' are the effective attributes.
In the first row of Figure~\ref{fig:E2}, where ``sex'' is the protected attribute, we adjusted ``age'' and ``hours-per-week''. However, the effective attribute ``education-num'' still shows a relatively high TV distance between genders (0.07, the highest among all non-adjusted attributes). In the second row, where ``race'' is the protected attribute, ``education-num'' is adjusted, and the other effective attributes exhibit a low TV distance between races.

\begin{figure}[!htb]
\centering
\includegraphics[width=0.5\linewidth]{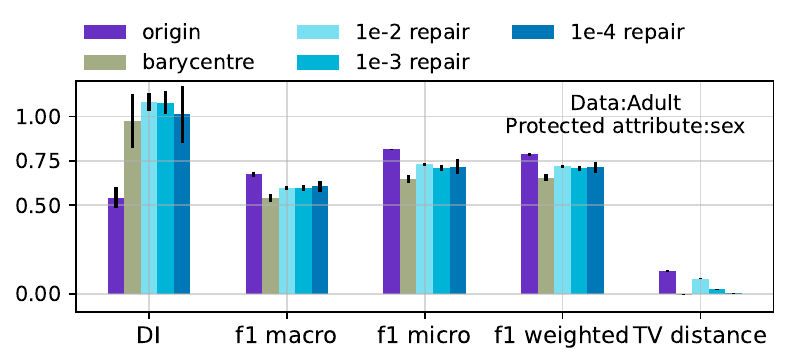}
\caption{\textcolor{black}{Prediction performance from the revised experiment in the first-row right column of Figure~\ref{fig:E2}, based on another 10-fold validation of the Adult dataset \citep{misc_adult_2}. The protected attribute is ``sex'', and the ineffective attribute ``hours-per-week'' (with a high TV distance) is removed, setting the $X$ attributes to ``age'' and ``education-num''. Bars indicate mean performance indices, with dark vertical lines representing the mean ± one standard deviation. Our partial repair schemes---``1e-2 repair'', ``1e-3 repair'', and ``1e-4 repair''---are shown in different shades of blue. Ideally, the disparate impact (``DI'') should be close to 1, while f1 scores should be maximised.}}
\label{fig:E3}
\end{figure}

To test this hypothesis, we modified our first-row experiments by removing ``hours-per-week'' (an ineffective attribute with high gender-wsie TV distance) and setting $X$ to ``age'' and ``education-num''.
The rest remain the same as before.
We conducted another 10-fold cross-validation to compare against ``origin'' and ``barycentre''. For our method, we set $\Theta$ to $(1e^{-2})\times\mathbb{1}$, $(1e^{-3})\times\mathbb{1}$ and $(1e^{-4})\times\mathbb{1}$ to visualise its effect, as shown in Figure~\ref{fig:E3}.
Initially, ``origin'' raised the ``DI'' from below 0.5 to slightly above, while maintaining the same accuracy as shown in Figure~\ref{fig:E2}, likely due to the removal of the ``hours-per-week'' attribute. 
Additionally, our ``1e-3 repair`` demonstrated improved mean performance of ``DI'', and lower f1 scores compared to Figure~\ref{fig:E2}. This might due to the change that all adjusted attributes are now effective. We also see slight improvement of ``DI'' performance in ``barycentre''.
The performances of ``1e-2 repair'', ``1e-3 repair'', were fairly similar, with the TV distance decreasing as $\Theta$ approaches the zero column. However, ``1e-4 repair'' began to show more instability.

\color{black}

\section{Conclusion}
We provided new bias repair schemes to mitigate disparate impact, via the technologies of optimal transport, without access to each datapoint's protected attribute. 
The limitation is that only one binary attribute is allowed.
We could further extend these schemes to cases where the protected attribute has multiple classes, or extend to the accelerated Dykstra’s algorithm in \cite{chai2022self}.

\acks{
We would like to acknowledge the contributions of Anthony Quinn and Robert Shorten, who suggested \citep{Quinn2023} 
to work on fairness repair without the protected attribute, following discussions within the AutoFair project as to the unavailability of the protected attributes in many settings. 
The original contribution of Anthony Quinn and Robert Shorten was to consider ``’research data'' \citep[Slide 23]{Quinn2023}, as in a regulatory sandbox,
as a means to repair data-sets with an unavailable protected attribute.
This insight has been instrumental in deriving Theorem \ref{pro:binary_condition}, which utilises marginals $P^{X_{s_0}},P^{X_{s_1}}$ of the 
general population. 
We would also like to extend our sincere thanks to Robert Shorten for his subsequent invaluable help, including his suggestion to focus on attribute distribution information via unbiased sampling from a wider population and to test the feasibility of our formulation using numerical experiments, 
as well as his proofreading of the manuscript.
This work has been done in parallel with the work of \cite{elzayn2023estimating}, whose pre-print appeared in arxiv on October 2nd, 2023, 
two weeks ahead of the present pre-print. 
This work has received funding from the European Union’s Horizon Europe research and innovation programme under grant agreement No. 101070568. This work was also supported by Innovate UK under the Horizon Europe Guarantee; UKRI Reference Number: 10040569 (Human-Compatible Artificial Intelligence with Guarantees (AutoFair)). 
}

\appendix
\section{Proof of Lemma~\ref{lem:tx|us}}
\label{app:lem:tx|us}
The right equation follows from the definition of coupling, as in Equation~\eqref{equ:Pi-define}. We only prove the left equation.
Since the projection $\mathcal{T}$ from $X$ to $\td{X}$ is irrelevant to $S$, the following holds
\begin{equation}
\pover{\td{X}=j}{X=i,S=s}=\pover{\td{X}=j}{X=i}, \forall i\in\supp{X_s}.
\end{equation}
Using Bayes theorem, for $i\in\supp{X_s}$:
\begin{align*}
\p{X=i,\td{X}=j,S=s}&=\pover{\td{X}=j}{X=i,S=s}\p{X=i,S=s}\\
&=\pover{\td{X}=j}{X=i}\p{X=i,S=s}\\
&=\p{X=i,\td{X}=j}\frac{\p{X=i,S=s}}{\p{X=i}}\\
&=\p{X=i,\td{X}=j}\pover{S=s}{X=i}.
\end{align*}
Further, 
\begin{align*}
\pover{\td{X}=j}{S=s}&=\frac{\p{\td{X}=j,S=s}}{\p{S=s}}=\frac{\sum_{i\in\supp{X_s}}\p{X=i,\td{X}=j,S=s}}{\p{S=s}}\\
&=\frac{\sum_{i\in\supp{X_s}}\p{X=i,\td{X}=j}\pover{S=s}{X=i}}{\p{S=s}}\\
&=\sum_{i\in\supp{X_s}}\p{X=i,\td{X}=j}\frac{\pover{S=s}{X=i}}{\p{S=s}}\\
&=\sum_{i\in\supp{X_s}}\p{X=i,\td{X}=j}\frac{\pover{X=i}{S=s}}{\p{X=i}}.
\end{align*}
We obtain that for all $j\in\supp{\td{X}}$, $s\in\supp{S}$:
\begin{equation*}
\pover{\td{X}=j}{S=s}=\sum_{i\in\supp{X_s}}\p{X=i,\td{X}=j}\frac{\pover{X=i}{S=s}}{\p{X=i}}.\label{equ:tx_from_coupling}
\end{equation*}
Note that $\pover{X=i}{S=s}$ is not defined on $\supp{X}\setminus\supp{X_s}$.
If we set the undefined conditional probability be $\pover{X=i}{S=s}=0$ for $i\in\supp{X}\setminus\supp{X_s}$, and rewrite the equation above in matrix form, we complete the proof.

\section{Proof of Theorem~\ref{pro:binary_condition}}
\label{app:pro:binary_condition}
Total repair in Definition~\ref{def:total-repair} requires that for $ j\in\supp{\td{X}}$, the difference between $\pover{\td{X}=j}{s_0}$ and $\pover{\td{X}=j}{s_1}$ is zero. 
$\supp{X}$ could be divided into three disjoint subsets: $\supp{X_{s_0}}\cap\supp{X_{s_1}}$, $\supp{X_{s_0}}\setminus\supp{X_{s_1}}$ and $\supp{X_{s_1}}\setminus\supp{X_{s_0}}$. Let $\bb{X}^{0\wedge 1}$, $\bb{X}^{0-1}$ and $\bb{X}^{1-0}$ denote these three subsets.
Using Lemma~\ref{lem:tx|us}, we deduce that
\begin{align*}
&\pover{\td{X}=j}{s_0}-\pover{\td{X}=j}{s_1}\\
&=\sum_{i\in\supp{X_{s_0}}}\p{X=i,\td{X}=j}\frac{\pover{X=i}{S=s_0}}{\p{X=i}}-\sum_{i\in\supp{X_{s_1}}}\p{X=i,\td{X}=j}\frac{\pover{X=i}{S=s_1}}{\p{X=i}}\\
&=\sum_{i\in\bb{X}^{0\wedge 1}}\p{X=i,\td{X}=j}\frac{\pover{X=i}{S=s_0}}{\p{X=i}}+\sum_{i\in\bb{X}^{0-1}}\p{X=i,\td{X}=j}\frac{\pover{X=i}{S=s_0}}{\p{X=i}}\\
&\quad -\sum_{i\in\bb{X}^{0\wedge 1}}\p{X=i,\td{X}=j}\frac{\pover{X=i}{S=s_1}}{\p{X=i}}-\sum_{i\in\bb{X}^{1-0}}\p{X=i,\td{X}=j}\frac{\pover{X=i}{S=s_1}}{\p{X=i}}\\
&=\sum_{i\in\bb{X}^{0\wedge 1}}\p{X=i,\td{X}=j}\frac{\pover{X=i}{S=s_0}-\pover{X=i}{S=s_1}}{\p{X=i}}\\
&\quad +\sum_{i\in\bb{X}^{0-1}}\p{X=i,\td{X}=j}\frac{\pover{X=i}{S=s_0}}{\p{X=i}}-\sum_{i\in\bb{X}^{1-0}}\p{X=i,\td{X}=j}\frac{\pover{X=i}{S=s_1}}{\p{X=i}}.
\end{align*}
Hence, if we want to achieve total repair, the coupling $\gamma$ must satisfy for $j\in\supp{\td{X}}$
\begin{equation}
\begin{split}
&0=\sum_{i\in\bb{X}^{0\wedge 1}}\p{X=i,\td{X}=j}\frac{\pover{X=i}{S=s_0}-\pover{X=i}{S=s_1}}{\p{X=i}}\\
&+\sum_{i\in\bb{X}^{0-1}}\p{X=i,\td{X}=j}\frac{\pover{X=i}{S=s_0}}{\p{X=i}}-\sum_{i\in\bb{X}^{1-0}}\p{X=i,\td{X}=j}\frac{\pover{X=i}{S=s_1}}{\p{X=i}}.
\end{split}
\label{equ:coupling-condition-binary}
\end{equation}
Note that $\pover{X=i}{S=s}$ is not defined on $\supp{X}\setminus\supp{X_s}$.
Now, set $\pover{X=i}{S=s_0}=0$ for $i\in\bb{X}^{1-0}$ and $\pover{X=i}{S=s_1}=0$ for $i\in\bb{X}^{0-1}$, such that for all $i\in\supp{X}$, it holds $\p{X=i}>0$ and $\pover{X=i}{S=s}\geq 0$ for $x\in\supp{X},s\in\supp{S}$.
Equation~\eqref{equ:coupling-condition-binary} can be simplified into 
\begin{equation}
\sum_{i\in\supp{X}}\p{X=i,\td{X}=j}\frac{\pover{X=i}{S=s_0}-\pover{X=i}{S=s_1}}{\p{X=i}}=0, \forall  j\in\supp{\td{X}}.
\label{equ:coupling-condition-binary-simple}
\end{equation}
Rewrite it into the matrix form. Using the definition of the vector $V$, Equation~\eqref{equ:coupling-condition-binary-simple} is further simplified into
\begin{equation}
\gamma^{\tr} V=\mathbb{0}.
\end{equation}

\section{Proof of properties in Remark~\ref{rem:v-property}}
\label{app:rem:v-property}
Since we have defined the conditional probability be $\pover{X=i}{S=s}=0$ for $i\in\supp{X}\setminus\supp{X_s}$, in Appendix~\ref{app:lem:tx|us} or~\ref{app:pro:binary_condition},
we can observe that $\sum_{i\in\supp{X}}P^{x_{s_0}}=\sum_{i\in\supp{X}}P^{x_{s_1}}=1$.

The property (i) derives from the following:
\begin{equation*}
(P^X)^{\tr}V=\sum_{i\in\supp{X}}P^X_i V_i= \sum_{i\in\supp{X}}P^X_i \frac{P^{X_{s_0}}_i-P^{X_{s_1}}_i}{P^X_i}=\sum_{i\in\supp{X}}P^{X_{s_0}}_i-P^{X_{s_1}}_i=0. 
\end{equation*}

\textcolor{black}{
The property (ii) is proved via contradiction. 
By definition, if $V_i=0$, then $P^{X_0}_i=P^{X_1}_i$.
We define the set of non-zero entries of $V$ as $\overline{\supp{X}}:=\{i\in\supp{X}\mid V_i\neq 0\}$, and let $c:=\sum_{i\notin\overline{\supp{X}}}P^{X_0}_i=\sum_{i\notin\overline{\supp{X}}}P^{X_1}_i$. If $V_i>0$ for $i\in\overline{\supp{X}}$, such that $P^{X_0}_i-P^{X_1}_i>0$ for $i\in\overline{\supp{X}}$, then $\sum_{i\in\supp{X}}(P^{X_0}_i-P^{X_1}_i)=\sum_{i\in\supp{X}}P^{X_0}_i-\sum_{i\in\supp{X}} P^{X_1}_i=(1-c)-(1-c)>0$. This contradiction shows that non-zero entries of $V$ cannot be all positive. The contradiction for all negative is similar.}

The property (iii) derives from:
\begin{equation*}
\|V\|_1\leq\|P^{X_{s_0}}-P^{X_{s_1}}\|_1 \|\frac{1}{P^X}\|_1\leq (\|P^{X_{s_0}}\|_1+\|P^{X_{s_1}}\|_1 )\|\frac{1}{P^X}\|_1 =2\|\frac{1}{P^X}\|_1,
\end{equation*}
where the minus and division operators are element-wise. The first inequality uses H{\``o}lder's inequality, and the second inequality uses triangle inequality. The equality uses the fact that vectors $P^{X_{s_0}},P^{X_{s_1}}$ are from the set $\Sigma_N$. Hence, if the norm $\|\frac{1}{P^X}\|_1$ is finite, vector $V$ has finite $l_1$ norm, such that there are not infinite entries in $V$.

\section{Proof of Lemma~\ref{lem:S-wise TV}}
\label{app:lem:S-wise TV}
 Using the TV distance in Definition~\ref{def:tvdistance}, we can deduce that
\begin{align}
\tv{P^{\td{X}_{s_0}}}{P^{\td{X}_{s_1}}}&=\frac{1}{2}\|P^{\td{X}_{s_0}}-P^{\td{X}_{s_1}} \|_1=\frac{1}{2}\left\|\gamma^{\tr}\left(\frac{P^{X_{s_0}}-P^{X_{s_1}}}{P^{X}}\right) \right\|_1=\frac{1}{2}\left\|\gamma^{\tr}V\right\|_1,\label{equ:tv-equivalent}
\end{align}
where the second equation follows from Lemma~\ref{lem:tx|us} such that $P^{\td{X}_{s}}=\frac{P^{X_{s}}}{P^X}$, with the division operator being element-wise.
The third equation follows from the definition of vector $V$ in Equation~\eqref{equ:V-def}.
\color{black}
Given the assumption that $V$ has finite $l_1$ norm.  
As in Definition~\ref{def:partialrepair}, $\Theta$-repair implies that there exists a non-negative vector $\Theta$ with the following holds 
\begin{equation}
-\Theta_j\leq \sum_{i\in\supp{X}}\gamma_{i,j}V_i\leq\Theta_j,\quad \forall j\in\supp{\td{X}},
\end{equation}
such that the vector $\gamma^{\tr}V$ is bounded by $\Theta$, i.e., $\|\gamma^{\tr} V\|_1\leq \|\Theta\|_1$.
Using Equation~\eqref{equ:tv-equivalent}, we know that $\tv{P^{\td{X}_{s_0}}}{P^{\td{X}_{s_1}}}=\left\|\gamma^{\tr}V\right\|_1/2\leq\|\Theta\|_1/2$.
\color{black}

\section{Feasibility proof of Lemma~\ref{lem:existence}}
\label{app:lem:existence}
The subset of all admissible couplings is
\begin{equation*}
\Pi_{\Theta }(P^{X},P^{\td{X}}):=\{\gamma\in\Sigma^2_N\mid \gamma\mathbb{1}=P^{X}, \gamma^{\tr}\mathbb{1}=P^{\td{X}},\left|\gamma^{\tr} V\right|_j\leq\Theta_j,\forall j\in\supp{\td{X}}\},
\end{equation*}
where $\left|\gamma^{\tr} V\right|_j:=\left|\sum_{i\in\supp{X}}\gamma_{i,j}V_i\right|$. Recall the definition of feasible couplings in Equation~\eqref{equ:Pi-define}, we can equivalently write 
\begin{equation*}
\Pi_{\Theta }(P^{X},P^{\td{X}})=\Pi(P^{X},P^{\td{X}})\cap\left\{ {\left|\gamma^{\tr} V\right|_j\leq\Theta_j,\forall j\in\supp{\td{X}}} \right\},
\end{equation*}
such that $\Pi_{\Theta }(P^{X},P^{\td{X}})\subseteq\Pi(P^{X},P^{\td{X}})$.
Now for any two non-negative vectors $\Theta',\Theta\in\bb{R}^{N}_+$, with their entries satisfying $0\leq\Theta'_j\leq\Theta_j$ for all $j\in\supp{\td{X}}$, the following holds
\begin{equation*}
\left\{ {\left|\gamma^{\tr} V\right|_j\leq\Theta'_j,\forall j\in\supp{\td{X}}} \right\} \subseteq \left\{ {\left|\gamma^{\tr} V\right|_j\leq\Theta_j,\forall j\in\supp{\td{X}}} \right\}.
\end{equation*}
Hence, we can deduce that
\begin{equation}
\begin{gathered}
\Pi(P^{X},P^{\td{X}})\cap\left\{ {\left|\gamma^{\tr} V\right|_j\leq\Theta'_j,\forall j\in\supp{\td{X}}} \right\} \subseteq \Pi(P^{X},P^{\td{X}})\cap\left\{ {\left|\gamma^{\tr} V\right|_j\leq\Theta_j,\forall j\in\supp{\td{X}}} \right\}\\
\iff  \Pi_{\Theta'}(P^{X},P^{\td{X}})\subseteq\Pi_{\Theta}(P^{X},P^{\td{X}}).    
\end{gathered}
\label{equ:theta-sequence}
\end{equation}
The set $\Pi_{\Theta}(P^{X},P^{\td{X}})$ becomes $\Pi(P^{X},P^{\td{X}})$ when $\Theta$ approaches infinity, and it becomes $\Pi_{\mathbb{0}}(P^{X},P^{\td{X}})$ when $\Theta$ is zero. Along with Equation~\eqref{equ:theta-sequence}, we conclude the following sequence
\begin{equation}
\Pi_{\mathbb{0}}(P^{X},P^{\td{X}})\subseteq\Pi_{\Theta'}(P^{X},P^{\td{X}})\subseteq\Pi_{\Theta}(P^{X},P^{\td{X}})\subseteq\Pi(P^{X},P^{\td{X}}).
\label{equ:pi-sequence}
\end{equation}

Next, we show that the set $\Pi_{\mathbb{0}}(P^{X},P^{\td{X}})$ is non-empty.
Remark~2.13 in \cite{peyre2019computational} states that $\Pi(P^{X},P^{\td{X}})$ is non-empty with an example of $P^{X}\otimes P^{\td{X}}\in\Pi(P^{X},P^{\td{X}})$, where $\otimes$ is the outer product. To verify this, we define a coupling $\gamma$ with its entries being $\gamma_{i,j}:=P^{X}_i P^{\td{X}}_j$. 
Using the fact $\sum_{i\in\supp{X}}P^{X}_i=\sum_{j\in\supp{\td{X}}}P^{\td{X}}_j=1$, we observe
\begin{align*}
\sum_{j\in\supp{\td{X}}}\gamma_{i,j}=\sum_{j\in\supp{\td{X}}}P^{X}_i P^{\td{X}}_j=P^{X}_i\sum_{j\in\supp{\td{X}}}P^{\td{X}}_j=P^{X}_i, \forall i\in\supp{X};\\
\sum_{i\in\supp{X}}\gamma_{i,j}=\sum_{i\in\supp{X}}P^{X}_i P^{\td{X}}_j=P^{\td{X}}_j\sum_{i\in\supp{X}}P^{X}_i =P^{\td{X}}_j,\forall j\in\supp{\td{X}};
\end{align*}
such that this coupling does belong to $\Pi(P^{X},P^{\td{X}})$.
Using Property (i) in Remark~\ref{rem:v-property}, we further observe 
\begin{equation*}
\sum_{i\in\supp{X}}\gamma_{i,j} V_i=P^{\td{X}}_j \sum_{i\in\supp{X}}P^{X}_i V_i=P^{\td{X}}_j\times \left((P^X)^{\tr} V\right)=0,\forall j\in\supp{\td{X}},
\end{equation*}
such that this coupling also belongs to $\Pi_{\mathbb{0}}(P^{X},P^{\td{X}})$.
Since we found one coupling in the set $\Pi_{\mathbb{0}}(P^{X},P^{\td{X}})$, this set is non-empty.
Using Equation~\eqref{equ:pi-sequence}, we know that for arbitrary non-negative $\Theta\in\bb{R}^N_+$, the feasible set of our formulation in Equation~\eqref{equ:our-formulation-1} is non-empty.

\section{Proof of Lemma~\ref{lem:subgradient-optimality}}
\label{app:lem:subgradient-optimality}
The proof uses subgradient optimality conditions that can be found in Proposition~5.4.7 in \cite{bertsekas2009convex}.
We start from restating the constrained KL projection as an unconstrained problem:
\begin{equation}
\prox^{KL}_{\mathcal{C}}(\bar{\gamma})=\arg\min_{\gamma\in\mathcal{C}}\kl{\gamma}{\bar{\gamma}}=\arg\min_{\gamma\in\bb{R}^{N\times N}_+}\kl{\gamma}{\bar{\gamma}} + \iota_{\mathcal{C}}(\gamma).
\end{equation}
The indicator function of a convex set $\mathcal{C}$ is a convex function, see Section~2 in \cite{ekeland1999convex}. Further, $\iota_{\mathcal{C}}$ is lower semi-continuous if and only if $\mathcal{C}$ is closed. Since $\mathcal{C}$ is a closed convex set (e.g., affine hyperplanes, closed half-spaces bounded by affine hyperplanes), $\iota_{\mathcal{C}}$ is a convex, lower semi-continuous coercive function.
Also, $\kl{\cdot}{\bar{\gamma}}$ is a strictly convex and coercive function.
This operator is uniquely defined by strict convexity.

As $\kl{\cdot}{\bar{\gamma}}$ is a Gateaux differential function with derivative $\nabla\kl{\gamma}{\bar{\gamma}}=\partial\kl{\gamma}{\bar{\gamma}}=\log(\gamma/\bar{\gamma})$.
Then the necessary and sufficient condition for $\gamma^*$ being the minimiser is
\begin{equation}
\mathbb{0}\in\partial\left(\kl{\gamma^*}{\bar{\gamma}}+\iota_{\mathcal{C}}(\gamma^*)\right)=\log\left(\frac{\gamma^*}{\bar{\gamma}}\right)+\partial\iota_{\mathcal{C}}(\gamma^*),
\end{equation}
where $\partial\iota_{\mathcal{C}}(\gamma^*)$ is the set of all subgradients at $\gamma^*$, and $\mathbb{0}$ is an $N\times N$ zero matrix.
The equation comes from $\dom\; \kl{\cdot}{\bar{\gamma}}\cap \textrm{dom}\;\iota_{\mathcal{C}}\neq\emptyset$.
To prove this optimality condition, define a function $J(\gamma)=\kl{\gamma}{\bar{\gamma}}+\iota_{\mathcal{C}}(\gamma)$ on $\bb{R}^{N\times N}_+$.
Due to $J(\gamma)\geq J(\gamma^*)$ for all $\gamma\in\bb{R}^{N\times N}_+$, if $\mathbb{0}\in\partial J(\gamma^*)$, then the definition of subgradients in Equation~\eqref{equ:subgradient-define} implies $J(\gamma)\geq J(\gamma^*)+\langle\mathbb{0},(\gamma-\gamma^*)\rangle$ for all $\gamma\in\bb{R}^{N\times N}_+$.

\section{Proof of Lemma~\ref{lem:c_1&c_2}}
\label{app:lem:c_1&c_2}
The proof is based on Appendix~A of \cite{peyre2015entropic}. 
Let $\gamma^*=\prox^{KL}_{\mathcal{C}_{\ell}}(\bar{\gamma})$.
\color{black}
For the first case, i.e., $\ell=1$, note that 
\begin{equation}
\iota_{\mathcal{C}_1}(\gamma)=\iota_{P^X}(\gamma\mathbb{1}),  \quad \partial\iota_{\mathcal{C}_1}(\gamma) =\partial\iota_{P^X}(P)\mathbb{1}^{\tr},
\end{equation}
where $P=\gamma\mathbb{1}$, and the second equation uses subgradient calculus.
$\iota_{P^X}(P)$ is zero if $P=P^X$, and infinity otherwise.
\color{black}
According to Lemma~\ref{lem:subgradient-optimality}, there exists a subgradient $u\in\partial\iota_{P^X}(P^*)\subset\bb{R}^{N\times 1}$, where $P^*=\gamma^*\mathbb{1}$, that satisfies
\begin{equation}
\mathbb{0}=\log\left(\frac{\gamma^*}{\bar{\gamma}}\right)+u\mathbb{1}^{\tr} \Rightarrow \gamma^*=\diag{\exp{(-u)}}\bar{\gamma}.
\end{equation}
Then, plug in $\gamma^*\mathbb{1}=P^X$, we have
\begin{equation}
\gamma^*\mathbb{1}=\diag{\exp(-u)}\bar{\gamma}\mathbb{1}=P^X \Rightarrow
\gamma^*=\diag{\frac{P^{X}}{\bar{\gamma}\mathbb{1}}}\bar{\gamma}.
\end{equation}
For the second case, i.e., $\ell=2$, the optimality condition becomes
\begin{equation}
\mathbb{0}=\log\left(\frac{\gamma^*}{\bar{\gamma}}\right)+\mathbb{1}\nu \Rightarrow \gamma^*=\bar{\gamma}\diag{\exp{(-\nu)}},
\end{equation}
where $\nu\in\partial\iota_{P^{\td{X}}}(\mathbb{1}^{\tr}\gamma^*)\subset\bb{R}^{1\times N}$.
Then, using $(\gamma^*)^{\tr}\mathbb{1}=P^{\td{X}}$, we have
\begin{equation}
(\gamma^*)^{\tr}\mathbb{1}=\diag{\exp(-\nu)}\bar{\gamma}^{\tr}\mathbb{1}=P^{\td{X}} \Rightarrow
\gamma^*=\bar{\gamma}\diag{\frac{P^{\td{X}}}{\bar{\gamma}^{\tr}\mathbb{1}}}.
\end{equation}

\section{Proof of Lemma~\ref{lem:prox_c3}}
\label{app:lem:prox_c3}
Let $\gamma^*=\prox^{KL}_{\mathcal{C}_3}(\bar{\gamma})$ and $[\bar{\gamma}^{\tr}V]_j$ be the $j^{th}$ entry of the vector $\bar{\gamma}^{\tr}V$.
\color{black}
If the entries of $\bar{\gamma}$ are irrelevant to or satisfy the constraint in $\mathcal{C}_3$, they should carry over to $\gamma^*$ in order to minimise the term $\kl{\gamma^*}{\bar{\gamma}}$ in the objective:
\begin{equation}
\gamma^*_{i,j}:=\bar{\gamma}_{i,j},\textrm{ if } i\notin\overline{\supp{X}} \textrm{ or } [\bar{\gamma}^{\tr}V]_j\in[-\Theta_j,\Theta_j].
\end{equation}
\color{black}
Therefore, we only discuss the entries $(i,j)$ satisfying both $i\in\overline{\supp{X}}$ and $j\in\{\supp{\td{X}}\mid[\bar{\gamma}^{\tr}V]_j\notin[-\Theta_j,\Theta_j]\}$. The existence of such entries implies that $\bar{\gamma}\notin\mathcal{C}_3$.
According to Lemma~\ref{lem:subgradient-optimality}, there exists a subgradient $\nu\in\partial\iota_{\Theta}(P^*)\subset\bb{R}^{1\times N}$, with $P^*=V^{\tr}\gamma^*$, such that
\begin{equation}
\mathbb{0}=\log\left(\frac{\gamma^*}{\bar{\gamma}}\right)+V{\nu} \Rightarrow \gamma^*=\bar{\gamma}\odot\exp{(-V{\nu})},
\end{equation}
where $\odot$ is element-wise product. 
The indicator function $\iota_{\Theta}(P)$ is 0 if $-\Theta\leq P\leq\Theta$ and infinity otherwise.
Recalling the definition of subgradient in Equation~\eqref{equ:subgradient-define}, we have
\begin{equation}
\langle \nu, (\gamma-\gamma^*)^{\tr}V\rangle + \iota_{\Theta}(V^{\tr}\gamma^*)\leq \iota_{\Theta}(V^{\tr}\gamma), \forall \gamma\in\bb{R}^{N\times N}_+.\label{equ:subgradient-character}
\end{equation}
By definition, we have $\gamma^*\in\mathcal{C}_3$.
Then, the right part of Equation~\eqref{equ:subgradient-character} is $\iota_{\Theta}(V^{\tr}\gamma)=0$ if the arbitrary matrix $\gamma\in\mathcal{C}_3$, and $\iota_{\Theta}(V^{\tr}\gamma)=\infty$ if the arbitrary matrix $\gamma\notin\mathcal{C}_3$. 
We equivalently simplify the condition in Equation~\eqref{equ:subgradient-character} into 
\begin{equation*}
\langle \nu, (\gamma-\gamma^*)^{\tr}V\rangle\leq 0, \forall \gamma\in\mathcal{C}_3.
\end{equation*}
Now, we gather these conditions found so far: 
\begin{align}
\boxed{
\langle \nu, (\gamma-\gamma^*)^{\tr}V\rangle\leq 0, \forall \gamma\in\mathcal{C}_3;\quad
\gamma^*=\bar{\gamma}\odot\exp{(-V{\nu})};\quad
\gamma^*\in\mathcal{C}_3.}
\label{equ:c_3_conditions}
\end{align}

\begin{itemize}
\item 
If $\Theta=\mathbb{0}$, 
given the third condition in Equation~\eqref{equ:c_3_conditions}, 
the first condition holds for any $\nu$ because $\gamma^{\tr}V={(\gamma^*)}^{\tr} V=\mathbb{0}$. The last two conditions require that $\nu$ needs to satisfy
\begin{equation*}
{(\gamma^*)}^{\tr} V=[\bar{\gamma}\odot\exp{(-V{\nu})} ]^{\tr}V =\mathbb{0}.
\end{equation*}
\item
If $\Theta\neq\mathbb{0}$, to make sure there exists a subgradient $\nu$ that satisfies the first condition, for any arbitrary matrix $\gamma\in\mathcal{C}_3$,
the optimal solution $\gamma^*$ must be on the edge of $\mathcal{C}_3$:
\begin{equation}
[{(\gamma^*)}^{\tr} V]_j=\Theta_j \textrm{ or } -\Theta_j, \forall j\in\supp{\td{X}}.\label{equ:nu-condition-0}
\end{equation}
It is obvious that the sign of $\nu_j$ should be the same as the sign of $[{(\gamma^*)}^{\tr} V]_j$.

Recall that we only focus on the case of $\bar{\gamma}\notin\mathcal{C}_3$, such that there are some $j\in\supp{\td{X}}$:
\begin{equation*}
\sum^N_{i=1} \bar{\gamma}_{i,j} V_i > \Theta_j \textrm{ or } \sum^N_{i=1} \bar{\gamma}_{i,j} V_i <-\Theta_j.
\end{equation*}
As the exponential function in the second condition is positive, i.e., $\exp{(-V_i\nu_j)}>0$, we cannot change the sign of $[{(\gamma^*)}^{\tr} V]_j$ by adjusting $\nu$. Hence, the last two conditions in Equation~\eqref{equ:c_3_conditions} imply that the subgradient $\nu$ should satisfy
\begin{equation}
[{(\gamma^*)}^{\tr} V]_j=\sum^N_{i=1} \bar{\gamma}_{i,j}\exp{(-V_i\nu_j)}V_i=
\begin{cases}
\Theta_j & \textrm{if } \sum^N_{i=1} \bar{\gamma}_{i,j} V_i > \Theta_j\\
-\Theta_j & \textrm{if } \sum^N_{i=1} \bar{\gamma}_{i,j} V_i < -\Theta_j
\end{cases}
,\forall j\in\supp{\td{X}}.
\label{equ:nu_j-app}
\end{equation}
\end{itemize}

Once the subgradient $\nu$ is found, plug in the second condition in Equation~\eqref{equ:c_3_conditions}, we find the optimal solution $\gamma^*$.

Note that the first and third conditions in Equation~\eqref{equ:c_3_conditions} guarantees the existence of subgradient $\nu$, i.e., the subdifferential is non-empty, while we also need (at least) one $\nu$ in the subdifferential to satisfy the second condition (i.e., the optimality condition).
The existence of such subgradient $\nu$ that satisfies Equation~\eqref{equ:nu-condition-0} or~\eqref{equ:nu_j-app} is discussed in Appendix~\ref{app:rem:nu_j}.

\section{Existence and finiteness of $\nu_j$}
\label{app:rem:nu_j}
Recall the definition of $\nu_j$ in the case of $\Theta\neq\mathbb{0}$ in Equation~\eqref{equ:nu_j-app}, with the case of $\Theta=\mathbb{0}$ being a special case.
Existence of such $\nu_j$ is equivalent to existence of (at least) one root of the function $F(x):=\sum_{i\in\overline{\supp{X}}}\bar{\gamma}_{i,j}V_i\exp{(-V_i x)}+c$, where the constant $c=\pm\Theta_j$.
Regardless of the value of this constant $c$, the first derivative of this function is $\nabla F(x)=-\sum_{i\in\overline{\supp{X}}}\bar{\gamma}_{i,j}(V_i)^2\exp{(-V_i x)}\leq 0$,
such that $F(x)$ is non-increasing.

We assume that the vector $V\neq\mathbb{0}$, because no coupling is needed otherwise.
Assumption~\ref{ass:finite} ensures that the vector $V$ has finite $l_1$-norm, such that none of its entries are infinity. 
We only discuss these entries $j\in\supp{\td{X}}$ where $[\bar{\gamma}^{\tr}V]_j > \Theta_j$ or $[\bar{\gamma}^{\tr}V]_j <- \Theta_j$, because we do not need to compute $\nu_j$ for other $j$ where $-\Theta\leq[\bar{\gamma}^{\tr}V]_j\leq\Theta_j$, as mentioned in Appendix~\ref{app:lem:prox_c3}.
Next, we discuss the  existence of roots (i.e., $\nu_j$).
\begin{itemize}
\item
If $[\bar{\gamma}^{\tr}V]_j > \Theta_j$, let $F(x):=\sum_{i\in\overline{\supp{X}}}\bar{\gamma}_{i,j}V_i\exp{(-V_i x)}-\Theta_j$, such that $\nu_j$ is the root of $F(x)$.
We first observe that
\begin{equation*}
[\bar{\gamma}^{\tr}V]_j > \Theta_j \Rightarrow F(0)=\sum_{i\in\overline{\supp{X}}}\bar{\gamma}_{i,j}V_i-\Theta_j> 0.
\end{equation*}
Since $F(x)$ is non-increasing, the root of $F(x)$ must be positive if it exists. Note that $F(x)$ includes the sum of the term $\bar{\gamma}_{i,j}V_i\exp{(-V_i x)}$. Let $x\rightarrow+\infty$, we observe that
\begin{equation*}
\lim_{x\to+\infty}\bar{\gamma}_{i,j}V_i\exp{(-V_i x)}=
\begin{cases}
0 & \textrm{ if } V_i>0 \textrm{ or }\bar{\gamma}_{i,j}=0\\
-\infty & \textrm{ if } V_i<0 \textrm{ and } \bar{\gamma}_{i,j}>0
\end{cases}
,\forall i\in\overline{\supp{X}}.
\end{equation*}

\color{black}
Therefore,
\begin{equation*}
\lim_{x\to+\infty}F(x)=
\begin{cases}
-\Theta_j & \textrm{ if } \nexists i\in\overline{\supp{X}}: V_i<0 \textrm{ and } \bar{\gamma}_{i,j}>0\\
-\infty & \textrm{ otherwise }
\end{cases}.
\end{equation*}
However, if the case of $\nexists i\in\overline{\supp{X}}: V_i<0$ and $\bar{\gamma}_{i,j}>0$ happens, then it holds that 
\begin{equation}
\nexists i\in\overline{\supp{X}}:\bar{\gamma}_{i,j} V_i < 0 \Rightarrow \forall i\in\overline{\supp{X}}:\bar{\gamma}_{i,j} V_i \geq 0.\label{equ:root1-app}
\end{equation}
By definition, $\bar{\gamma}_{i,j}\geq 0$, and $V_i\neq 0$ for $i\in\overline{\supp{X}}$. Equation~\eqref{equ:root1-app} implies that $V_i\geq 0$ for all $i\in\overline{\supp{X}}$, which is in conflict with property (ii) of Remark~\ref{rem:v-property}. Therefore, the case of $\nexists i\in\overline{\supp{X}}: V_i<0$ and $\bar{\gamma}_{i,j}>0$ is impossible to happen, such that $\lim_{x\to+\infty}F(x)=-\infty$.
Then, using the fact that $F(0)>0$ and $\lim_{x\to+\infty}F(x)<0$, the root of function $F(x)$ must exist, such that $\nu_j$ exists and falls within $(0,+\infty)$. 
\color{black}

\item If $[\bar{\gamma}^{\tr}V]_j < -\Theta_j$, let $F(x):=\sum_{i\in\overline{\supp{X}}}\bar{\gamma}_{i,j}V_i\exp{(-V_i x)}+\Theta_j$, such that $\nu_j$ is the root. 
Similarly, we observe that
\begin{equation*}
[\bar{\gamma}^{\tr}V]_j < -\Theta_j \Rightarrow F(0)=\sum_{i\in\overline{\supp{X}}}\bar{\gamma}_{i,j}V_i+\Theta_j< 0.
\end{equation*}
As $F(x)$ is non-increasing, the root of $F(x)$ is negative if it exists. Let $x\rightarrow-\infty$, the summand of the function $F(x)$ goes to
\begin{equation*}
\lim_{x\to-\infty}\bar{\gamma}_{i,j}V_i\exp{(-V_i x)}=
\begin{cases}
0 & \textrm{ if } V_i<0 \textrm{ or } \bar{\gamma}_{i,j}=0\\
+\infty & \textrm{ if } V_i>0 \textrm{ and }\bar{\gamma}_{i,j}>0
\end{cases}
,\forall i\in\overline{\supp{X}}.
\end{equation*}

\color{black}
Therefore,
\begin{equation*}
\lim_{x\to-\infty}F(x)=
\begin{cases}
\Theta_j & \textrm{ if } \nexists i\in\overline{\supp{X}}: V_i>0 \textrm{ and }\bar{\gamma}_{i,j}>0\\
+\infty & \textrm{ otherwise }
\end{cases}.
\end{equation*}
Similarly, the case of $\nexists i\in\overline{\supp{X}}: V_i>0$ and $\bar{\gamma}_{i,j}>0$ is impossible to happen due to the contradiction with property (ii) of Remark~\ref{rem:v-property}. Hence, $\lim_{x\to-\infty}F(x)=+\infty$.
Then using the fact that $F(0)<0$, and $\lim_{x\to-\infty}F(x)>0$, the root of function $F(x)$ must exist, such that $\nu_j$ exists and falls within $(-\infty,0)$.
\color{black}
\end{itemize}



We have shown in both cases, that the root $\nu_j$ exists and has finite value.
This guarantees that our computation in Appendix~\ref{app:lem:prox_c3} can find the solution of $\prox^{KL}_{\mathcal{C}_3}(\bar{\gamma})$ as long as the root-finding algorithm converges to the root.

\section{Convergence proof of Dykstra's algorithm}
\label{app:Dykstra}

This section heavily depends on \cite{bauschke2000dykstras} and Section~30.2 in \cite{bauschke2017correction}.

Assume $E$ is some Euclidean space with inner product $\langle\cdot,\cdot\rangle$ and induced norm $\|\cdot\|$. Without loss of generality, we assume there are two closed convex sets $\mathcal{C}_1,\mathcal{C}_2$ in $E$ with $\mathcal{C}:=\mathcal{C}_1\cap\mathcal{C}_2\neq\emptyset$. 
Note that the assumption of $\mathcal{C}\neq\emptyset$ in Dykstra's algorithm is satisfied due to our feasibility results in Lemma~\ref{lem:existence}. 

\begin{definition}[Bregman projections]\label{def:app-Bregman}
The general Bregman projection of $\xi$ onto $\mathcal{C}$ respect to a Legendre function $\varphi$ on $E$ and the corresponding Bregman distance 
\begin{equation*}
\begin{split}
D_{\varphi}: E\times \interior(\dom \varphi)&\to [0,+\infty]:\\
(\gamma,\xi)&\mapsto \varphi(\gamma)-\varphi(\xi)-\langle\nabla\varphi(\xi),\gamma-\xi\rangle,
\end{split}
\end{equation*}
is to solve the optimisation problem:
\begin{equation*}
\prox^{\varphi}_{\mathcal{C}}(\xi):=\min_{\gamma\in\mathcal{C}\cap \dom \varphi} D_{\varphi}(\gamma,\xi).
\end{equation*}
Note that when $\varphi(\gamma):=\sum_{i,j}\gamma_{i,j}\log \gamma_{i,j}$, the general Bregman divergence is KL divergence, as defined in Equation~\eqref{equ:kl-define}.    
\end{definition}

Given an arbitrary point $\xi$ in $E$.
Since only the KL projection is considered in this paper, we use $\prox_{\mathcal{C}}(\xi)$ in place of $\prox^{\varphi}_{\mathcal{C}}(\xi)$, to denote the point in $\mathcal{C}$ that is the nearest to $\xi$, and we assume $\mathcal{C}_k:=\mathcal{C}_k\cap\interior(\dom\varphi)$, $k=1,2$ for ease of notation.
The Dykstra's algorithm in \cite{boyle1986method,bauschke2017correction} sets $\gamma_0:=\xi,q_0=q_{-1}=q_{-2}:=0$, and then iteratively conducts the following sequences for $k\geq 1$:
\begin{equation}
\gamma_k:=\prox_{\mathcal{C}_k}(\gamma_{k-1}+q_{k-2}),\quad q_k:=\gamma_{k-1}+q_{k-2}-\gamma_k, \label{equ:sequence-define}
\end{equation}
where the ``$2$'' comes from the number of convex sets considered in this problem, and $\prox_{\mathcal{C}_k}$ is the Bregman projections onto the convex set $\mathcal{C}_{(k\mod 2)}$. 
The sequence $\{\gamma_k\}_{k\geq 1}$ converges strongly to $\prox_{\mathcal{C}}(\xi)$, \cite{boyle1986method}.

\begin{lemma}[Theorem 3.16 in \cite{bauschke2017correction}]\label{lem:heorem 3.16}
When $\mathcal{C}$ is a non-empty closed convex subset of $E$, for every $x,\gamma_k\in E$,
\begin{equation*}
\gamma_{k}=\prox_{\mathcal{C}}(x) \Leftrightarrow \gamma_{k}\in\mathcal{C} \textrm{ and } \langle c-\gamma_k,x-\gamma_k\rangle\leq 0, \forall c\in\mathcal{C}.
\end{equation*}
\end{lemma}

\begin{lemma}\label{lem:s_1<=0}
For any $k>0$, $\gamma_k=\prox_{\mathcal{C}_k}(\gamma_{k-1}+q_{k-2})$ and then for any $c\in\mathcal{C}_k$, the following holds
\begin{equation*}
\gamma_k\in\mathcal{C}_{k},\quad\langle c-\gamma_{k},q_k\rangle\leq 0.
\end{equation*}
\begin{proof}
From Lemma~\ref{lem:heorem 3.16} and Equation~\eqref{equ:sequence-define}, 
as $\mathcal{C}_k$ is a non-empty closed convex subset of $E$, we deduce that for $(\gamma_{k-1}+q_{k-2}),\gamma_k\in E$, the following holds
\begin{equation*}
\gamma_{k}=\prox_{\mathcal{C}_k}(\gamma_{k-1}+q_{k-2}) \Leftrightarrow \gamma_{k}\in\mathcal{C}_k \textrm{ and } \langle c-\gamma_k,\gamma_{k-1}+q_{k-2}-\gamma_k\rangle=\langle c-\gamma_k,q_k\rangle\leq 0.
\end{equation*}
\end{proof}
\end{lemma}

\begin{lemma}\label{lem:seq-bounded}
The sequence $\{\gamma_k\}_{k\geq 1}$ is bounded and $\gamma_{k-1}-\gamma_{k}\rightarrow 0$, where the arrow ``$\rightarrow$'' refers to strong convergence.
\begin{proof}
From Equation~\eqref{equ:sequence-define}, for any $k\geq 1$, the following holds
\begin{equation}
\gamma_{k-1}-\gamma_{k}=q_k - q_{k-2}.\label{equ:seq-property-2}
\end{equation}
Let $c\in\mathcal{C}$, we deduce that
\begin{align*}
\|\gamma_{k}-c\|^2&=\|\gamma_{k+1}-c\|^2 + \|\gamma_{k}-\gamma_{k+1}\|^2+2\langle\gamma_{k+1}-c,\gamma_{k}-\gamma_{k+1}\rangle\\
&=\|\gamma_{k+1}-c\|^2 + \|\gamma_{k}-\gamma_{k+1}\|^2+2\langle\gamma_{k+1}-c,q_{k+1}-q_{k-1}\rangle,
\end{align*}
where the second equality used Equation~\eqref{equ:seq-property-2}.
Then we rewrite the equation above into
\begin{equation*}
\|\gamma_{k+1}-c\|^2-\|\gamma_{k}-c\|^2 = -\|\gamma_{k}-\gamma_{k+1}\|^2
-2\langle\gamma_{k+1}-c,q_{k+1}\rangle -2\langle \gamma_{k-1}-\gamma_{k+1},q_{k-1}\rangle
+2\langle\gamma_{k-1}-c,q_{k-1}\rangle.
\end{equation*}
Using induction, we deduce that for $l\geq n$, the following holds
\begin{align*}
&\|\gamma_{l}-c\|^2-\|\gamma_{n}-c\|^2\\&=\sum_{k=n}^{l-1}\Big(\|\gamma_{k+1}-c\|^2-\|\gamma_{k}-c\|^2\Big)\\
&=-\sum_{k=n}^{l-1}\Big( \|\gamma_{k}-\gamma_{k+1}\|^2 +2\langle \gamma_{k-1}-\gamma_{k+1},q_{k-1}\rangle \Big)+\sum_{k=n}^{l-1}\Big(2\langle\gamma_{k-1}-c,q_{k-1}\rangle-2\langle\gamma_{k+1}-c,q_{k+1}\rangle \Big)\\
&=-\sum_{k=n}^{l-1}\Big( \|\gamma_{k}-\gamma_{k+1}\|^2 +2\langle \gamma_{k-1}-\gamma_{k+1},q_{k-1}\rangle \Big)+
2\sum_{k=n-1}^{n}\langle\gamma_{k-1}-c,q_{k-1}\rangle
-2\sum_{k=l}^{l+1}\langle\gamma_{k+1}-c,q_{k+1}\rangle.\\
\end{align*}
Now, set $n=0$, such that $2\sum_{k=n-1}^{n}\langle\gamma_{k-1}-c,q_{k-1}\rangle=0$ by $q_{-1}=q_{-2}=0$. We have 
\begin{equation}
\|\gamma_{l}-c\|^2-\|\xi-c\|^2=-\sum_{k=0}^{l-1}\Big( \|\gamma_{k}-\gamma_{k+1}\|^2 +2\langle \gamma_{k-1}-\gamma_{k+1},q_{k-1}\rangle \Big)
-2\sum_{k=l}^{l+1}\langle\gamma_{k+1}-c,q_{k+1}\rangle.\label{equ:bounded-difference}
\end{equation}
By definition in Equation~\eqref{equ:sequence-define}, $\gamma_{k-1},\gamma_{k+1},c\in\mathcal{C}_{(k-1\mod 2)}$. Using Lemma~\ref{lem:s_1<=0}, we have
\begin{equation*}
\langle\gamma_{k-1}-\gamma_{k+1},q_{k-1}\rangle\geq 0,\langle\gamma_{k+1}-c,q_{k+1}\rangle\geq 0.
\end{equation*}
Go back to Equation~\eqref{equ:bounded-difference}, we can deduce that
\begin{equation*}
\|\gamma_{l}-c\|^2 \leq\|\xi-c\|^2,\forall l\geq 1,
\end{equation*}
such that the sequence $\{\gamma_{k}\}_{k\geq 1}$ is bounded and $\lim_{l}\sum_{k=0}^{l-1} \|\gamma_{k}-\gamma_{k+1}\|^2 < +\infty$. We can now conclude that the sequence of partial sums $\{\sum_{k=0}^{l-1} \|\gamma_{k}-\gamma_{k+1}\|^2\}_{l\geq 1}$ is a monotone non-negative real sequence with bounded limit, then the difference $\gamma_{k}-\gamma_{k+1}$ tend to zero.
\end{proof}
\end{lemma}

\begin{lemma}[Lemma~30.6 in \cite{bauschke2017correction}]\label{lem:Lemma 30.6}
Suppose $\{p_k\}_{k\geq 1}$ is a sequence of non-negative reals with $\sum_{k}p^2_k<+\infty$. Let $\sigma_n:=\sum_{k=1}^{n}p_k$. Then $\liminf_{n}\sigma_n(\sigma_n-\sigma_{n-m-1})=0$, for an arbitrary $1\leq m\leq n$.
\end{lemma}

\begin{lemma}
\begin{equation*}
\liminf_{n}|\langle\gamma_{n-1}-\gamma_{n},q_k\rangle|=0.
\end{equation*}
\begin{proof}
Since $q_k=(q_{k}-q_{k-2})+(q_{k-2}-q_{k-4})+\cdots+(q_{(k\mod 2)}-q_{(k\mod 2)-2})$, using triangle inequality for norm, we have
\begin{equation}
\|q_{n-1}\|\leq\|q_n\|+\|q_{n-1}\|\leq \sum_{k=1}^{n}\|q_k-q_{k-2}\|=\sum_{k=1}^{n}\|\gamma_{k-1}-\gamma_{k}\|,\label{equ:q-sequ-inequ}
\end{equation}
where the equality utilises Equation~\eqref{equ:seq-property-2}. The case of $k=0$ is omitted because $q_0,q_{-1},q_{-2}$ are zero. Further,
\begin{align*}
|\langle\gamma_{n-1}-\gamma_{n},q_{n-1}\rangle|&\leq\|\gamma_{n-1}-\gamma_{n}\|\|q_{n-1}\|\\
&\leq \|\gamma_{n-1}-\gamma_{n}\|
\left(\sum_{k=1}^{n}\|\gamma_{k-1}-\gamma_{k}\|\right).
\end{align*}
The first inequality uses Cauchy-Schwarz inequality. The second one uses Equation~\eqref{equ:q-sequ-inequ}. Then, using Lemma~\ref{lem:seq-bounded} and~\ref{lem:Lemma 30.6}, we complete the proof.
\end{proof}
\end{lemma}

\begin{lemma}\label{lem:s_1+s_2}
For any $c\in\mathcal{C}$,
\begin{equation*}
\langle c-\gamma_{n},\xi-\gamma_{n}\rangle= \sum_{k=n-1}^{n}\langle c-\gamma_{k},q_k\rangle+\sum_{k=n-1}^{n}\langle\gamma_k-\gamma_{n},q_k\rangle
\end{equation*}
\begin{proof}
Applying induction to Equation~\eqref{equ:seq-property-2}, we observe that
\begin{equation}
\xi-\gamma_n=\sum_{k=1}^{n}\gamma_{k-1}-\gamma_{k}=\sum_{k=1}^{n} q_k -q_{k-2}=\sum_{k=n-1}^{n} q_k.\label{equ:seq-property-3}
\end{equation}
Hence, 
\begin{align*}
\langle c-\gamma_{n},\xi-\gamma_{n}\rangle &=\sum_{k=n-1}^{n}\langle c-\gamma_{n}, q_k\rangle=\sum_{k=n-1}^{n}\langle c-\gamma_{k},q_k\rangle+\sum_{k=n-1}^{n}\langle\gamma_k-\gamma_{n},q_k\rangle.
\end{align*}
\end{proof}
\end{lemma}

\begin{lemma}[Lemma~2.42 in \cite{bauschke2017correction}]\label{lem:lemma 2.42}
If $\gamma_{k_n}$ converges weakly to $c^*\in\mathcal{C}$, then
\begin{equation*}
\|c^*\|\leq \liminf_{n} \|\gamma_{k_n}\|.
\end{equation*}
\begin{proof}
By Cauchy-Schwarz inequality, 
\begin{equation*}
\|c^*\|^2= \lim_{n} |\langle\gamma_{k_n},c^*\rangle|\leq \liminf_{n} \|\gamma_{k_n}\|\|c^*\|.
\end{equation*}
\end{proof}
\end{lemma}

\begin{theorem}[\cite{boyle1986method}]
We can find a subsequence $\{k_n\}_{n\geq 1}$ of $\{k\}_{k\geq 1}$ with
\begin{equation}
\begin{split}
\lim_{n}|\langle\gamma_{k_n-1}-\gamma_{k_n},q_k\rangle|=0.
\end{split}\label{equ:subsequence}
\end{equation}
This subsequence converges weakly to $P_{\mathcal{C}}(\xi)\in\mathcal{C}$. 
\begin{proof}
From the definition of the subsequence, $\{\gamma_{k_n}\}_{n\geq 1}$ converges weakly to some $c^*\in\interior(\dom \varphi)$ and $\lim_{n}\|\gamma_{k_n}\|$ exists. From the definition of the sequence in Equation~\ref{equ:sequence-define}, we know that $\gamma_{k_n}\in\mathcal{C}_{(k_n\mod 2)}$ for $k_n>0$. Using Lemma~\ref{lem:seq-bounded}, we know that $c^*\in\mathcal{C}$.

Further, we can deduce from Lemma~\ref{lem:s_1<=0} and \ref{lem:s_1+s_2} that for any $c\in\mathcal{C}$: 
\begin{align*}
\limsup_{n}\langle c-\gamma_{k_n},\xi-\gamma_{k_n}\rangle
&=\limsup_{n}\left(\sum_{l=k_n-1}^{k_n}\langle c-\gamma_{l},q_{l}\rangle+\sum_{l=k_n-1}^{k_n}\langle\gamma_{l}-\gamma_{k_n},q_l\rangle\right) \\
&=\limsup_{n}\sum_{l=k_n-1}^{k_n}\langle c-\gamma_{l},q_l\rangle\leq 0.
\end{align*}

Lemma~\ref{lem:lemma 2.42} shows that if $\gamma_{k_n}$ converges weakly to $c^*$, for any $c\in\mathcal{C}$,
\begin{equation}
\begin{split}
\langle c-c^*,\xi-c^*\rangle\leq\limsup_{n}\langle c-\gamma_{k_n},\xi-\gamma_{k_n}\rangle\leq 0.
\end{split}\label{equ:subsequence-converges}
\end{equation}
Combining Equation~\eqref{equ:subsequence-converges} and Lemma~\ref{lem:heorem 3.16}, we conclude that $c^*=P_{\mathcal{C}}(\xi)$, such that this subsequence converges weakly to $P_{\mathcal{C}}(\xi)$. 
\end{proof}
\end{theorem}

For the case of two convex sets $\mathcal{C}_1,\mathcal{C}_2$, the fact that the subsequence converges weakly to $P_{\mathcal{C}}(\xi)$ and $\lim_{n}\|\gamma_{n}-\gamma_{n+1}\|=0$ by Lemma~\ref{lem:seq-bounded} are sufficient to show that the whole sequence converges strongly to $P_{\mathcal{C}}(\xi)$.
For the strong convergence proof of more than two convex sets, we refer to Section~30.2 in \cite{bauschke2017correction} and \cite{bauschke2000dykstras} for more details.

\bibliography{ref}

\end{document}